\pdfoutput=1

\documentclass[11pt]{article}

\usepackage[]{ACL2023}

\usepackage{times}
\usepackage{latexsym}
\usepackage{multirow}
\usepackage{graphicx}
\usepackage{caption}
\usepackage{subcaption}

\usepackage[T1]{fontenc}

\usepackage[utf8]{inputenc}

\usepackage{microtype}

\usepackage{inconsolata}
\usepackage{kotex}
\usepackage[flushleft]{threeparttable} 
\usepackage{makecell}
\usepackage{graphicx}
\usepackage{booktabs}
\usepackage{multicol}
\usepackage{makecell}
\usepackage{amsmath}
\usepackage{geometry}
\usepackage{array}
\usepackage{tabularx}
\usepackage{amsmath}
\usepackage{longtable}

\newcommand{\ours}{\textsc{kbl}}
\newcommand{\Ours}{\textsc{Korean Benchmark for Legal language understanding}}

\newcommand{\kblk}{\textsc{knowledge}}
\newcommand{\kblr}{\textsc{reasoning}}
\newcommand{\kbll}{\textsc{bar exam}}

\newcommand{\kcommonMistakeqa}{\textsc{mstke}}

\newcommand{\kcommonMistakeqaR}{\textsc{mstke$_r$}}
\newcommand{\kconceptqa}{\textsc{conc}}
\newcommand{\kcomponentqa}{\textsc{comp}}
\newcommand{\kstatuteMatching}{\textsc{stat}}
\newcommand{\kstatuteQueryMatching}{\textsc{stat$_q$}}
\newcommand{\kstatuteHallucination}{\textsc{hall}} 

\newcommand{\rcausal}{\textsc{causal}}
\newcommand{\rcontradict}{\textsc{cons}}
\newcommand{\rcaseq}{\textsc{rel$_q$}}
\newcommand{\rcasep}{\textsc{rel$_p$}}

\newcommand{\bcriminal}{\textsc{criminal}}
\newcommand{\bcivil}{\textsc{civil}}
\newcommand{\bpublic}{\textsc{public}}
\newcommand{\bresponsibility}{\textsc{resp}}

\title{Developing a Pragmatic Benchmark for Assessing Korean Legal Language Understanding in Large Language Models}

\author{Yeeun Kim$^\sigma$\\ \texttt{keeun04@uos.ac.kr} 
        \And Young Rok Choi$^\lambda$ \\ \texttt{yrchoi@lbox.kr}
        \And Eunkyung Choi$^\sigma$ \\ \texttt{rmarud202@uos.ac.kr}
        \AND Jinhwan Choi$^\lambda$ \\ \texttt{jinhwanchoi@lbox.kr}
        \And Hai Jin Park$^{\eta}$ \\ \texttt{haijinpark@hanyang.ac.kr}
        \And Wonseok Hwang$^{\sigma,\lambda,}$\thanks{\quad Corresponding author. Also reachable via email at \texttt{wonseok.hwang@uos.ac.kr}} \\ \texttt{wonseok.hwang@lbox.kr}
        \AND $^\sigma$University of Seoul \quad \quad $^\lambda$LBox \quad \quad $^\eta$Hanyang University
        }

\begin{document}
\maketitle
\begin{abstract}
Large language models (LLMs) have demonstrated remarkable performance in the legal domain, with GPT-4 even passing the Uniform Bar Exam in the U.S. However their efficacy remains limited for non-standardized tasks and tasks in languages other than English. This underscores the need for careful evaluation of LLMs within each legal system before application.
Here, we introduce \ours, a benchmark for assessing the Korean legal language understanding of LLMs, consisting of (1) 7 legal knowledge tasks (510 examples), (2) 4 legal reasoning tasks (288 examples), and (3) the Korean bar exam (4 domains, 53 tasks, 2,510 examples). First two datasets were developed in close collaboration with lawyers to evaluate LLMs in practical scenarios in a certified manner. Furthermore, considering legal practitioners' frequent use of extensive legal documents for research, we assess LLMs in both a closed book setting, where they rely solely on internal knowledge, and a retrieval-augmented generation (RAG) setting, using a corpus of Korean statutes and precedents. The results indicate substantial room and opportunities for improvement.

\end{abstract}

\section{Introduction}
Large Language Models (LLMs) show remarkable performance across various tasks. For instance, GPT-4 has passeed the Uniform Bar Exam~\cite{martinez2023reeval_bar_exam_gpt4,openai2023gpt4}.
However, their performance is still limited beyond the standardized bar exam questions \cite{guha2023legalbench,dahl2024largelegalfiction,magesh2024lhallucinationfree,kang2023-chatgpt-irac,bernsohn-2024-eacl-legallens,blairstanek2023gpt3performstatutoryreasoning}, particularly in handling legal tasks in languages other than English \cite{fei2023lawbench,zhong2023agieval}.
This implies it is necessary to thoroughly evaluate LLMs before applying them to specific legal tasks and systems.

Previous works have developed legal benchmarks for evaluating LLMs primarily in English~\cite{guha2023legalbench} and Chinese~\cite{fei2023lawbench,dai2023laiw,collam2024} mostly focusing on a closed book setting where LLMs answer questions based solely on the knowledge stored in their parameters. 
Considering that legal practitioners often rely on previous legal documents such as statutes or precedents to make decision, this setting is underestimate LLMs' capabilities in a practical scenarios.

Here we introduce \ours\footnote{\Ours}, a benchmark dedicated to assessing LLMs' capability in understanding Korean legal language.
\ours\ consists of (1) 7 legal knowledge tasks (510 examples), (2) 4 legal reasoning tasks (288 examples), and (3) the Korean bar exam (4 domains, 53 tasks, 2,510 examples). 
We evaluate LLMs' under two settings to better reflect their real-world usage of LLMs. In collaboration with legal professionals, we focus on the desgin and quality assurance of the tasks moving beyond solely standardized licensure-style questions.
Specifically, we assess LLMs in scenarios where (1) they cannot access external knowledge, and (2) they can use retrievers to search for related documents. For the second setting, we prepare a Korean statute corpus consisting of 11k currently active articles and municipal ordiances and rules, and utilized a corpus of 150k Korean precedents released from a previous study \cite{hwang2022lboxopen}.

The results indicates that the performance of arguably the most powerful LLMs, such as GPT-4 and Claude-3, remains limited in handling Korean legal tasks. In an open book setting, their accuracy improves by up to +8.6\%, but the overall performance still varies depending on the type of corpus and the LLMs used. This suggests significant potential for improvement not only in the LLMs themselves but also in the methods of document retrieval and their integration.

In summary, our contributions are as follows.
\begin{itemize}
    \item We have developed the first pragmatic and verified benchmark for Korean legal understanding tasks.
    \item We evaluate LLMs not only in a closed book setting but also in a open book setting, where they can utilize two critical legal resources: Korean statutes and precedents.
    \item We compare various LLMs and provide the detailed analysis.
\end{itemize}
Our datasets, the corpus for RAG, and the evaluation code will be released to the community under a CC BY-NC license\footnote{\url{https://github.com/lbox-kr/kbl}}.

\section{Related Work}
Many legal AI datasets has been created contributing to NLP community \cite{paul2022lesicin,kapoor2022facl_hindi_legal_corpus,yao2022FACLlevenEventDetection,glaser2021nllu_german_court_summarization,niklaus2021nllu_swiss_ljp,chalkidis2022acl_lexglue,rossi2021verbcl,chalkidis2022facl_fairlex,louis2022acl_belgian_statutory_retrieval,rabelo2020coliee,henderson2022pileoflaw,chalkidis2023lexlms,niklaus2023multilegalpile}.
Here we review only a few works that are dedicated to build benchmark for evaluating LLMs under zero, or few-shot setting in legal domain.

\citet{guha2023legalbench} developed LegalBench which comprises 162 legal language understanding tasks organized according to six different types of legal reasoning based on the IRAC framework. However, their work is limited to tasks in English legal language understanding tasks. Additionally, the benchmark does not evaluate LLMs in a open-book setting, and LLMs must rely solely on their internal knowledge.

\citet{magesh2024lhallucinationfree} compared commercial RAG-based AI systems in the US legal domain using 202 examples on generative tasks and found that even the most competent system exhibited a 17\% hallucination rate through human evaluation.
In contrast, our research focuses on Korean legal AI tasks and emphasizes scenarios where automatic evaluations are possibles. 
RAG systems rely on various components including databases, search algorithms, and the backbone of LLMs. Given these complexities, it becomes 
 impractical to manually evaluate the performance of RAG systems every time a component changes. 
Therefore, we concentrate on addressing the challenges where automatic evaluation is feasible.

\citet{fei2023lawbench} developed LawBench that consists of 20 Chinese legal tasks. They categorized the tasks into three different levels Memorization, Understanding, and Applying based on Bloom’s taxonomy. \ours\ differs in that it handles Korean legal language understanding tasks and it evaluates LLMs not just in a closed-book setting but also in a open-book setting by introducing accompanying legal corpus.

\citet{hwang2022lboxopen} developed LBoxOpen, which includes two text classification tasks, two legal judgement prediction tasks, and one summarization task in the Korean legal domain. Each task contains 5k--50k examples for fine-tuning language models and evaluating performance. The datasets, except for the summarization dataset, were created by semi-automatically processing Korean precedents, while human-annotated examples were used for summarization. Additionally, they open-sourced 150k Korean precedents. \ours\ differs by focusing on (1) evaluating the zero-shot performance of (large) language models across (2) more diverse tasks (7 knowledge, 4 reasoning, 53 Korean bar exam), using (3) expert-annotated examples created spedifically for this study.

\citet{son2024kmmlu} developed KMMLU, a dataset similar to MMLU\cite{hendryckstest2021mmlu} tailored for Korean and cultural contexts. 
It includes diverse language understanding tasks along with four legal AI tasks named \texttt{criminal-law, law, patent}, and \texttt{taxation}, sourced from the Public Service Aptitude Test and various Korean license exams.\footnote{The specific origins of the datasets are not described in the paper.} Many examples pertain to licensure exams such as the Petitioner Police exam, the Real Estate Brokers exam, and the Occupational Safety and Health exam, or cover topics in social science theory and legal terminology. 
These tasks are generally less challenging than those found in the bar exam. 
In contrast, \ours\ consists of questions from the Korean bar exam and the newly created questions designed to assess comprehensive legal knowledge and in-depth reasoning capability relevant to the practice of law. 
Our work also differs in that it exclusively focus on the legal domains, and we have collaborated with legal professionals to develop a  pragmatic and verified benchmark. 
Additionally, we also assess LLMs in scenarios where they can utilize legal documents. 
We have also ensured that there is no overlap between the legal task examples in KMMLU and our dataset by conducting fuzzy matching between the two datasets, with the most significant matches displayed in Tables \ref{tab:comparison_kmmlu_korean_bar_civil}--\ref{tab:comparison_kmmlu_korean_bar_criminal} in Appendix.

In addition to KMMLU, several other benchmarks have been developed specifically to evaluate language models on Korean natural language processing tasks.
\citet{park2021klue} introduced KLUE for Korean Language Understanding tasks.
\citet{kim2022kobestkoreanbalancedevaluation} developed KoBEST, targeting tasks that require advanced Korean-specific knowledge.
\citet{son-etal-2024-haerae} created HAE-RAE benchmark, which assess Korean cultural knowledge.
\citet{park-etal-2024-open-ko-h5} developed Ko-H5 benchmark, consisting of Korean translated and reviewed datasets. Our work differs in that it focuses on evaluating LLMs on Korean legal language understanding tasks.

\section{Datasets}
Our benchamrk consists of 7 knowledge tasks, 4 reasoning tasks,\footnote{Here, the term ``reasoning'' refers to ``general NLP reasoning task in the legal domain''} and multiple-choice questions from the Korean bar exam. 
All tasks are structured as question-answering task where the model must {\it select} the correct answer for given questions, similar to  MMLU~\cite{hendryckstest2021mmlu}. 
The datasets were compiled using various sources, including Korean precedents, statutes, bar exams, legal QA datasets from Korea Legal Aid Corporation\footnote{\url{https://www.klac.or.kr/}}, etc.

The Korean legal system, rooted in civil law, incorporates complex historical and cultural aspects unique to Korea. Notably, GPT-4 has not passed the Korean bar exam, despite passing the US bar exam. This highlights that creating a Korean legal benchmark involves more than translating existing benchmarks; it requires developing a new system with a unique ontology.
To ensure the quality of the datasets, we developed the tasks in close corporation with legal professionals, and all the answers have been verified by 8 licensed lawyers.\footnote{For two case relevance QA datasets (\rcaseq, \rcasep), only a portion of examples were verified due to the difficulty of the tasks. For more details, see Section \ref{sec: reasoning_datasets}} The verification took 26 hours in total reflecting the difficulty (and the quality) of the tasks constructed. 
During the quality assurance process, we found freely available data, often created by individuals with semi-expertise, frequently include substantial amounts of errors (up to 21\% in our study), highlighting the importance of close collaboration with experts (Section \ref{sec: general_lessons} covers additional general lessons learned during the dataset creation process).
Representative examples from individual tasks are displayed in Tables \ref{tab:task_examples_1}, \ref{tab:task_examples_2}, \ref{tab:task_examples_3} in Appendix. 

\begingroup
\setlength{\tabcolsep}{2pt} 
\renewcommand{\arraystretch}{1} 
\begin{table*}[th!]
\scriptsize
  \caption{Data statistics. The GPT-4o tokenizer was used.}
  \label{tbl_data_stat}
  \centering
  \begin{threeparttable}
  \begin{tabular}{l|ccccccc|cccc|cccc}
    \toprule
    \multicolumn{1}{c}{Name} &
    \multicolumn{7}{c}{\makecell{\kblk}} &
    \multicolumn{3}{c}{\makecell{\kblr}} &
    \multicolumn{4}{c}{\makecell{\kbll$^\dagger$}} 
    \\
        
      & \kconceptqa 
      & \kcomponentqa
      & \kstatuteMatching 
      & \kstatuteQueryMatching 
      & \kstatuteHallucination
      & \kcommonMistakeqa
      & \kcommonMistakeqaR
      & \rcausal
      & \rcontradict
      & \rcaseq
      & \rcasep
      & \bcriminal
      & \bcivil 
      & \bpublic 
      & \bresponsibility
      \\
    \midrule
    n$_\text{samples}$
      & 100
      & 102
      & 100
      & 52
      & 75
      & 41
      & 40
      & 95
      & 91
      & 46
      & 56
      & 520
      & 910
      & 520
      & 560
      \\
    n$_\text{tok}$ 
      & 170
      & 208
      & 715
      & 194
      & 200
      & 113
      & 167
      & 1462
      & 211
      & 4,288
      & 8,858
      & 551
      & 495
      & 556 
      & 360
      \\
    \bottomrule
  \end{tabular}
  \begin{tablenotes}[]
  \item $\dagger$: The number of all examples; Criminal law 2012--2024, 40  examples per year; Civil law 2012--2024, 70 examples per year; Public law 2012--2024, 40 examples per year; Responsibility 2010--2023, 40 examples per year.
  \end{tablenotes}
  \end{threeparttable}
\end{table*}
\endgroup

\subsection{Legal Knowledge Tasks}

\paragraph{Legal Concept QA}
The legal concept QA dataset (\kconceptqa) comprises questions addressing complex legal concepts. 
For instance, from various types of suspension--such as suspension of indictment, suspension of sentence, suspension of execution, suspension of collection, suspension of announcement--a model needs to select the option that best fits a given definition.
The examples were crafted based on legal terminology reference documents from South Korean courts\footnote{\url{https://sldongbu.scourt.go.kr/word/new/WordList.work}}, definition of legal terms provided by the National Law Information Center\footnote{\url{https://www.law.go.kr/LSW/eng/engMain.do}}, and the Korea Legislation Research Institute\footnote{\url{https://www.klri.re.kr/kor/business/bizLawDicKeyword.do}}. Please See Section \ref{sec: appendix_legal_concept_qa} for further information.

\paragraph{Offense Component QA}
The offense component QA dataset (\kcomponentqa) comprises question and answer pairs that determine whether specific actions meet the actual elements of a criminal offense. The dataset was constructed using several sources: ``100 Questions and 100 Answers'' on Day-to-Day Laws\footnote{\url{https://www.easylaw.go.kr/CSP/OnhunqueansLstRetrieve.laf?onhunqnaAstSeq=88&onhunqueAstSeq=439}} provided by the Ministry of Justice, data crawled from a now-defunct law firm website\footnote{\url{https://github.com/haven-jeon/LegalQA?tab=readme-ov-file}}, cases from Korea Legal Aid Corporation, from statutory interpretations from the Ministry of Government\footnote{\url{https://www.moleg.go.kr/lawinfo/nwLwAnList.mo?mid=a10106020000}}. 
The questions were refined to be clear legal inquiries based on responses from actual legal consultation experts\footnote{\url{https://www.klac.or.kr/legalinfo/counsel.do}}.
Responses are binary, either ``Yes'' or ``No''. For example, one real client question inquires whether escaping prison due to unbearable harassment over private loans by falsely reporting to authorities could constitute the crime of false accusation.

To ensure the quality of the dataset, we verified all questions with a lawyer and found that 20\% of examples contained incorrect answers\footnote{The free legal counselings aim to make legal services accessible to socially disadvantaged groups while stating that the answers provided may not be entirely accurate, as they are based solely on the questions and facts presented by the client.}. The lawyer provided feedback on why the responses from the free legal counselings were incorrect\footnote{For example, one of the feedback is "It is not possible to determine a violation of the Personal Information Protection Act based solely on the provided facts and questions. However, if the question is changed to inquire about defamation, it can be determined that the offense components are met."}. The dataset was revised accordingly and we conducted a second round of verification with the lawyer who confirmed that all 102 revised responses were free of errors.

\paragraph{Statue Matching QA}
The statute matching QA dataset was constructed from statutes currently active in South Korea. 
To compile the raw data, we first extracted and counted the citations of articles from approximately 3 million Korean precedents. Based on these statistics, we randomly sampled 100 articles--about one-third from the 50 most cited articles, one-third from the 50 least cited articles, and one-third the entire range. We also excluded frequently cited statutes such as the Civil Acts, Civil Procedure Act, Criminal Acts, and Criminal Procedure Act from the top 50 to ensure diversity. This method guarantees the inclusion of articles from various legal domains. The resulting dataset comprises 100 statutes, includes articles from the Civil Act, Enforcement Decree of the Public Service Ethics Act, Public Official Election Act, Pharmacists Act, Environmental Preservation Act, etc.

Based on the extracted articles, we developed two datasets: \kstatuteMatching\ and \kstatuteQueryMatching.
\kstatuteMatching\ comprises 100 questions designed to evaluate whether a model can accurately match the content of a law to its specific statute number. For example, language models are tested on their ability to determine whether Article 36 of the Food Sanitation Act pertains to facility standards or the evaluation of food standards and specifications.
\kstatuteQueryMatching\ includes 50 questions that assess whether the language model can accurately identify the relevant statute for a given query. For instance, in response to the question ``Is the maximum fine 2 million won for failing to report and instead keeping a lost item found on the street?'', the language model must correctly identify the articles related to the misappropriation of lost property.

\paragraph{Hallucination QA}
The hallucination QA dataset is constructed from laws that often confused by the general public.
For example, many people are unaware that throwing an object at someone can constitute assault, even if it does not hit the target. The dataset also includes laws specific to Korea, such as the Kim Young Ran Law, which prohibits gifts of food and drink worth more than 30,000 won to public officials. 
First, 75 examples were collected similarly to the \kcomponentqa\ dataset with following criteria: the questions should be commonly mistaken and reviewed by a lawyer. Approximately, around 21\% (16 out of 75) were corrected based on the lawyer’s feedback. These corrected examples were used in the three sub tasks: statute hallucination QA (\kstatuteHallucination), common legal mistake QA (\kcommonMistakeqa), and common legal mistake QA with reasoning (\kcommonMistakeqaR).
 For given confusing legal questions, a model needs to select the correct answer where the answer set consists of: (1) (fictitious) statute and corresponding reasoning (\kstatuteHallucination), or (2) ``yes'', ``no'', or ``there is no correct answer'' (\kcommonMistakeqa). In case of \kcommonMistakeqaR, the answer also includes corresponding reasoning.

\subsection{Legal Reasoning Tasks} \label{sec: reasoning_datasets}
\paragraph{Causal Reasoning QA}
The causal reasoning QA dataset (\rcausal) is compiled from a series of verdicts from criminal trials involving physical harm leading to  injury or death. The examples are drawn from cases categorized as ``Death or Injury Resulting from Violence'' or ``Death Resulting from Bodily Injury''. We used precedents from these two criminal cases based on advice from a lawyer, as selected cases had relatively well-matched causal relations. 
For each given factual description and claims, the task is to assess whether the defendant's actions were the direct and decisive cause of the victim's injury or death. A guilty verdict implies that ``there is causal relationship'' between the defendant’s actions and the victim's injury or death, indicating that the event would not have occurred without the defendant’s involvement. Conversely, a not guilty verdict may indicate that other factors also contributed to the victim's injury or death, or that there was no no causal relationship between the defendant's actions and the outcome. These instances are classified under ``no causal relationship.''

In the first verification process, we shared ~5 specific examples with four legal professionals and refined the guidelines based on their inputs. After collecting a total of 100 examples, we conducted a second verification process, where another lawyer who did not participate in the first process, reviewed all 100 samples and provided feedback on errors. The error rate was 7\%, and these erroneous examples were either replaced or removed according to the lawyer's feedback, resulting in a final 95 samples.

\paragraph{Statement Consistency QA}
Statement Consistency QA (\rcontradict) is developed using the verdicts from criminal and civil trials.
These verdict documents often note inconsistencies in the statements of the victims, witnesses, and defendants. 
If a statement given at the scene, during prosecution, or in court is found inconsistent with other evidences, it may be officially recorded as such by the judge.
In this task,  a model is required to accurately determine whether two presented statements are consistent with each other. 
It's important to note that the ``consistency'' is judged based on a legal perspective. For example, hitting 5 times can be considered as legally consistent compared to hitting 7 times depending on the circumstances\footnote{Suwon District Court, Anyang Branch, Judgment dated November 20, 2020, Case No. 2020고합56 (수원 지방법원 안양지원 2020. 11. 20. 선고 2020고합56 판결)}. 
We followed a similar verification process with the \rcausal\ dataset. In the second verification process, approximately 9\% of the data was questioned by a lawyer and corresponding examples were deleted, resulting in a final 91 examples.
Note that both \rcausal\ and \rcontradict\ datasets were built from Korean precedents where the decision is carefully made by judges. Also, although the second verification processes were conducted by a single lawyer, the decisions were made by citing corresponding precedents, providing unified guidelines for the decision.

\paragraph{Case Relevance QA} 
The two Case Relevance QA datasets (\rcaseq, \rcasep) are constructed based on how judges cite previous cases.
In the precedents from the Supreme Court of Korea (referred to as $P_S$), judges often references prior cases ($P_R$) in two contexts: (1) to support their decision citing similar cases (``relevant''), or (2) to denote that the cases brought by the appellant do not pertain to the  current case (``not relevant''). 

For the first dataset, \rcaseq, a model needs to determine whether a given precedent supports the query.
Each example in this dataset includes: (1) $P_S$, the Supreme Court precedent (2) $P_R$, a referenced precedent, (3) $L_{S}$, a precedent from a lower court handling the same case with  $P_S$, and (4) $L_R$, a precedent from a lower court related to $P_R$. 
To construct the queries, we 
(1) manually extracted the facts and the appellant's arguments from $P_S$ and $L_{S}$, 
(2) generate initial queries using GPT-4, 
and (3) manually revised these queries. The target precedent is composed by extracting the facts, the appellant's claim, and the judge's opinion from $P_R$ and $L_R$. This dataset primarily includes examples where $P_R$ is not supportive.
We also created positive examples where $P_R$ is supportive, but they were not included here based on advice from legal professionals.
This decision was influenced by two factors: (1) the queries do not incorporate the judge's opinion, which is crucial for determining the relevance between legal cases, and (2) each precedent encompasses various legal judgements where ``supportiveness'' is an issue-specific determination, hence the query should align specifically with the relevant part of the precedent.

To address these issues, we developed the second dataset, \rcasep. In this task, a model must determine whether two given precedents address the same legal issue. To improve relevance assessment, the previous queries in \rcaseq\ change into the precedent including judge's opinions. 
Additionally, we incorporate 10 positive examples where the conclusion of $P_S$ is ``remand for retrial,'' and the judge accepts $P_R$, which is provided by the appellants to support their claim. 
We assume the decision to ``remand for retrial'' strongly indicates ``supportiveness.''

Note that the ``relevance'' between cases depends on various aspects such as users' specific goals \cite{vanOpijnen2017ai_and_law_relevance}. In this regards, previous studies in legal information retrieval often address this issue by considering all possible relevance simultaneously \cite{santosh2024ecthrpcr,hou2024clercdatasetlegalcase}. 
For instance in the COLIEE legal case retrieval tasks, if a case is cited in another case (regardless of whether two cases are supportive or not), the cited case is considered as relevant \cite{goebel2023coliee}. 
In our approach, we attempt to differentiate ``relevance'' into two categories based on lexical cues found in the precedents.

\subsection{Korean Bar Exam}
\paragraph{Multiple-choice questions}
The Korean Bar Exam is designed to evaluate legal knowledge and the capability to perform tasks essential for a lawyer. Administered at least once annually under the Ministry of Justice's supervision, the exam is divided into two main parts: multiple-choice questions and an essay-type written test. 
The multiple-choice section comprises 150 questions, divided among Public Law (\bpublic), Civil Law (\bcivil), and Criminal Law (\bcriminal),  with 40, 70, and 40 questions in each subject area, respectively. The Public Law section covers Constitutional Law and Administrative Law; The Civil Law section encompasses Civil Law, Commercial Law, and Civil Procedure Law; and the Criminal Law section includes Criminal Law and Criminal Procedure Law. 
We use the test held in 2012--2024. 
The essay-type written test covers specialized legal areas such as International Law, Labor Law, and Tax Law, in addition to Public Law, Civil Law, and Criminal Law.

We focus solely on the multiple-choice questions for Public Law, Civil Law, and Criminal Law. as the official answers for the essay-type test are not publicly available. Additionally, the multiple-choice section offers clear and definitive answers, providing an ideal playground to evaluate LLMs' under a RAG setting, where multiple components can influence performance.\footnote{The Korean bar exam is published annually by the Ministry of Justice of Korea under 'KOGL Type 1' license. The license permits both commercial and non-commercial use, and allows the creation of derivative works, including modifications, as long as the source is cited.}

\paragraph{Professional Responsibility}
The Professional Responsibility (\bresponsibility) examination is a test that conducted by the Ministry of Justice to assess the professional ethics required of lawyers. Held at least once a year, this examination comprises 40 multiple-choice questions. We use the tests conducted from 2010 to 2023 for our analysis.

\subsection{Corpus}
We utilize 150k Korean precedents released by \cite{hwang2022lboxopen} for the RAG experiments.
The corpus, processed using the gpt-4o tokenizer, contains 320M tokens. 
Additionally, we have developed a new Korean statute corpus compiled from active Korean statutes (법령) and municipal ordinances and rules (자치법규) as of Nov. 2023. The raw data for this was collected from LAW OPEN DATA\footnote{https://open.law.go.kr/LSO/main.do}, a resource maintained by the Korean government. This statute corpus consists of 220k articles, totaling 52M tokens, where each article is concatenated with the name of the act, again processed using the gpt-4o tokenizer.

\begingroup
\setlength{\tabcolsep}{0.8pt} 
\renewcommand{\arraystretch}{1} 
\begin{table*}[ht!]
\scriptsize
  \caption{Comparison of various models. The accuracies of individual tasks are shown for
  legal concept QA (\kconceptqa), offense component QA (\kcomponentqa), statute number and content matching QA (\kstatuteMatching), statute and query matching QA (\kstatuteQueryMatching), statute hallucination QA (\kstatuteHallucination), common legal mistake QA (\kcommonMistakeqa), common legal mistake QA with reasoning (\kcommonMistakeqaR), causal reasoning QA (\rcausal), statement consistency QA (\rcontradict), query and case relevance QA (\rcaseq), inter-case relevance QA (\rcasep), and the Korean bar exam (\kbll). The average accuracies for 7 knowledge tasks (AVG$_K$), 2 reasoning tasks (\rcausal, \rcontradict, AVG$_R$), and 3 bar exams (AVG$_B$) are shown. For the experiments with GPT where the model shows randomness even with temperature $\rightarrow$ 0 due to their internal algorithm that automatically raises the temperature until certain threshold satisfied, we repeat either two (the knowledge and the reasoning tasks) or three (the bar exam) times and show their mean values. \texttt{prec.} and \texttt{stat.} in the bottom 5 rows indicate the precedents and the statutes corpus respectively. Two corpora were used during the RAG experiments. \texttt{n/a} indicates the scores cannot be computed due the limitation in the context length. The scores for other open-source models are present in Table \ref{tbl_comp_appendix} in Appendix.
  }
  \label{tbl_comp}
  \centering
  \begin{threeparttable}
  \begin{tabular}{l|c|ccccccc|c|cccc|c|ccc}
    \toprule
    \multicolumn{1}{c}{Name} &
    \multicolumn{8}{c}{\makecell{\kblk}} &
    \multicolumn{5}{c}{\makecell{\kblr}} &
    \multicolumn{4}{c}{\makecell{\kbll\ 2024$^\dagger$}} 
    \\
      & AVG$_K$  
      & \kconceptqa 
      & \kcomponentqa
      & \kstatuteMatching 
      & \kstatuteQueryMatching 
      & \kstatuteHallucination
      & \kcommonMistakeqa
      & \kcommonMistakeqaR
      & AVG$_R$  
      & \rcausal
      & \rcontradict
      & \rcaseq
      & \rcasep
      & AVG$_B$  
      & \bcriminal
      & \bcivil 
      & \bpublic 
      \\
      \midrule

      Most frequent
      & 33 
      & 20 
      & 50 
      & 20 
      & 21 
      & 25 
      & 34 
      & 35 
      & 50
      & 50
      & 50
      & 50
      & 50
      & 24
      & 23 
      & 26 
      & 23 
      \\
      \midrule
    EEVE-10.8b$^{a}$ (solar)
    & 45.8
    & 42.0
      & 45.1
      & 23.0
      & 80.8
      & 41.3
      & 51.2
      & 37.5
      & 58.4 
      & 42.1
      & 74.7
      & n/a
      & n/a 
      & 17.7
      & 12.5
      & 15.7
      & 25.0
    \\
    KULLM3-10.7b$^{b}$ (solar)
    & 52.1
     & 52.0
     & 53.9
      & 23.0
      & 94.2
      & 42.7
      & 53.6
      & 45.0
      & 80.1 
      & 81.1
      & 79.1
      & n/a 
      & n/a 
      & 20.0
      & 20.0
      & 20.0
      & 20.0
      \\
    LG-Exaone-3.0-7.8b$^{c}$
    & 59.3
     & 83.0
     & 53.9
      & 29.0
      & 98.1
      & 44.0
      & 48.8
      & 57.5
      & 79.1
      & 79.0
      & 79.1
      & n/a 
      & n/a 
      & 20.8
      & 22.5
      & 20.0
      & 20.0
      \\
      Qwen2-7b$^{d}$
    & 55.3
     & 60.0
     & 54.9
      & 32.0
      & 96.2
      & 48.0
      & 51.2
      & 45.0
      & 79.2 
      & 73.7
      & 84.6
      & n/a 
      & n/a 
      & 28.1
      & 35.0
      & 24.3
      & 25.0
      \\
    Qwen2-72b$^{e}$
     & 60.2
     & 82.0
     & 48.0
      & 34.0
      & 98.1
      & 53.3
      & 48.8
      & 57.5
      & 86.7
      & 82.1 
      & 91.2 
      & n/a 
      & n/a 
      & 31.1
      & 22.5
      & 35.7
      & 35.0
      \\
    \midrule
    Claude-3-sonnet$^f$
      & 62.0
      & 82.0
      & 51.0
      & 36.0
      & 98.0
      & 65.0
      &  56.0
      &  60.0
      & 87.7 
      & 85.3
      & 90.1
      & 15.2
      & 66.1
      & 33.5
      & 27.5
      & 32.9
      & 40.0
      \\   
    Claude-3-opus$^g$
    & 67.0
    & 89.0
      & 55.0
      & 37.0
      & 100
      & 80.0
      & 61.0
      & 57.0
      & 87.2 
      & 85.3
      & 89.0
      & 45.7
      & 78.6
      & 41.0
      & 27.5
      & 42.9
      & 52.5
      \\
    Claude-3.5-sonnet$^h$
      & 70.0
      & 93.0
      & 62.8
      & 42.0
      & 100
      & 81.3
      & 58.5
      & 52.5
      & 89.3
      & 87.4
      & 91.2
      & 56.5
      & 75.0
      & 42.5
      & 37.5
      & 40.0
      & 50.0
      \\   
    GPT-3.5$^i$
    & 50.0
    & 58.0
      & 49.0
      & 26.0
      & 93.0
      & 45.0
      & 46.0
      & 46.0
      & 62.7 
      & 71.1 
      &  54.5 
      &  n/a 
      &  n/a 
      & 23.1
      & 15.0
      & 24.3
      & 30.0
      \\
      GPT-4o-mini$^j$
      & 64.6
      & 83.0
      & 55.9
      & 31.0
      & 98.1
      & 70.7
      & 61.0
      & 52.5
      & 85.5 
      & 84.2
      & 86.8
      & 67.4
      & 76.8
      & 29.6
      & 25.0
      & 31.4
      & 32.5
      \\
    GPT-4$^k$
      & 72.0
      & 95.0
      & 64.0
      & 49.0
      & 100
      & 77.0
      & 61.0
      & 65.0
      & 88.6 
      & 84.2 
      & 92.9 
      & 79.4 
      & 81.3 
      & 48.1
      & 39.2 
      & 46.7 
      & 58.3 
      \\
      \midrule 
      GPT-4 + prec.
      & 74.4
      & 95.5 
      & 71.6 
      & - 
      & 100
      & 75.4 
      & 61.0 
      & 68.8 
      & -
      & -
      & -
      & -
      & -
      & 50.9
      & 41.6  
      & 55.2  
      & 55.8 
      \\
     GPT-4 + stat.
      &  72.4
      &  95.0 
      &  62.8  
      & - 
      &  100 
      &  78.7 
      &  56.1 
      &  65.0 
      & -
      & -
      & -
      & -
      & -
      & 49.3
      & 46.7  
      & 48.6  
      & 52.5 
      \\
    GPT-4 + prec. + stat.
    & 75.3
      & 95.0 
      & 72.6 
      & - 
      &  100
      &  79.4 
      &  61.0 
      &  70.0 
      
      & -
      & -
      & -
      & -
      & -
      &  49.7
      &  46.7 
      &  50.0 
      &  52.5 
      \\
      \midrule 
      Claude-3-sonnet + prec. + stat.
      & 59.9
      & 67.0
      & 53.9
      & -
      & 84.6
      & 66.7
      & 56.1
      & 55.0
      & -
      & -
      & -
      & -
      & -
      & 18.8
      & 12.5 
      & 21.4 
      & 22.5 
      \\
      Claude-3-opus + prec. + stat.
    & 65.3
      & 87.0
      & 59.8
      & -
      & 98.1
      & 66.7
      & 46.3
      & 62.5
      & -
      & -
      & -
      & -
      & -
      & 32.1 
      & 32.5 
      & 21.4 
      & 42.5 
      \\
      Claude-3.5-sonnet + prec. + stat.
      & 70.1
      & 92.0
      & 56.9
      & -
      & 98.1
      & 82.7
      & 51.2
      & 67.5
      & -
      & -
      & -
      & -
      & -
      & 46.2
      & 50.0
      & 38.6
      & 50.0
      \\

    \bottomrule
  \end{tabular}
  \begin{tablenotes}[]
  \item $^\dagger$The scores for Bar exams 2012--2023 shown in Appendix.
  \item $^a$\texttt{yanolja/EEVE-Korean-10.8B-v1.0}~~~$^b$\texttt{nlpai-lab/KULLM3}~~~$^c$\texttt{EXAONE-3.0-7.8B-Instruct} \citep{research2024exaone3078binstruction}~~~$^d$\texttt{Qwen2-72B-Instruct-GPTQ-Int8}
  \item $^e$\texttt{Qwen/Qwen2-7B-Instruct}~~~$^f$\texttt{claude-3-sonnet-20240229}~~~$^g$\texttt{claude-3-opus-20240229}~~~$^h$\texttt{claude-3-5-sonnet-20240620}
  \item $^i$\texttt{gpt-3.5-turbo-0125}~ ~~$^j$\texttt{gpt-4o-2024-0513}~~~$^k$\texttt{gpt-4o-mini-2024-07-18}

  \end{tablenotes}
  \end{threeparttable}
\end{table*}
\endgroup

\section{Experiments}
\subsection{Evaluation}
We developed the evaluation code using the \texttt{lm-eval-harness} framework~\cite{gao2023eval-harness}. 
Each task is formulated as multiple-choice question, where the model must generate a label corresponding to the given questions and possible selections. The options are tagged with letters A through E, and the model is tasked with generating the letter that matches the ground truth.
For the evaluation of open-source LLMs (EEVE~\cite{kim2024eeve}, KULLM~\cite{kim2023kullm3}, and Qwen2~\cite{bai2023qwen}), we compare the average logit values by feeding individual (question, choice) pairs into the models. 
This approach is adopted because the performance of models tend to drop significantly when they directly generate labels.

\subsection{Retrieval Augmented Generation}
We constructed the BM25 retriever using the \texttt{pyserini} library~\cite{Lin_etal_SIGIR2021_Pyserini}. 
We first indexed the precedent, and the statute corpora using \texttt{pyserini.index.lucene} and \texttt{LuceneSearcher} with default settings.
To retrieve related documents,  the question is used as a query to the retriever. The number of documents retrieved is determined in the  following way; (1) identify the maximum number of documents that LLMs can process simultaneously, (2) progressively decrease the number of documents until the performance ceases to improve. This results in retrieving between 1--8 documents; the exact number will be specified in the accompanying code.
For the \kstatuteHallucination\ task, we employ individual (question, choice) pairs as queries to retrieve related documents since each answer include the name of a (fictitious) statute. This method ensures the relevance of retrieved documents to the specific query, enhancing the accuracy of the task performance.

\section{Results}

We evaluate four open-source LLMs and five commercial LLMs across 7 legal knowledge tasks, 4 legal reasoning tasks, and 3 bar exam tasks conducted in 2024 (Table \ref{tbl_comp}).

\paragraph{Open-source LLMs show performance comparable to GPT-3.5}
We first compare open-source LLMs and GPT-3.5. 
On the knowledge tasks, EEVE-10.7b, KULLM3-10.7b, Qwen2-7b, and GPT-3.5 score on average 45.8, 52.1, 55.3, and 50.0 respectively (rows 2--6, 10; column 1) indicating open-source LLMs achieve comparable or better performance than GPT-3.5.
Similarly, on two reasoning tasks--\rcausal, \rcontradict--all open source LLMs outperform GPT-3.5 except for EEVE-10.7b, which scores 4.3 points lower.
 
However on the Korean bar exam which requires complex legal reasoning and knowledge, all open-source LLMs and GPT-3.5 perform close to the baseline achievable by selecting the most frequent labels (4th column from the right, rows 1--6, 9). 
Notably, Qwen2-72b shows the highest performance, even exceeding GPT-3.5 by margins of +10.2, +24.0, and +8.0 one knowledge tasks, reasoning tasks, and the bar exam respectively (rows 6 vs 9). \footnote{Even Qwen2-7b shows most competent  performance compared to other open-source Korean LLMs of similar sizes, particularly in Bar examss. We propose two possible explanations for this. As previously reported \citet{chalkidis2020legalbert,hwang2022lboxopen}, training legal LLMs from scratch can be beneficial for solving difficult tasks. But, most strong open-source Korean LLMs are adapted from English-dedicated LLMs (Llama \cite{touvron2023llama,touvron2023llama2}, Solar \cite{kim-etal-2024-solar}, Mistral \cite{jiang2023mistral7b}) by further training with a Korean corpus. There may be knowledge transfer between multilingual corpus. Note that the Qwen series also achieves strong performance in the Chinese legal domain \cite{anonymouse2024collm}. Although the Korean legal system differs from Chinese legal systems, basic legal frameworks like IRAC may have corresponding notions and legal terms in each system. It is important to note that the details of the training corpus for many open-source LLMs are not fully disclosed, making it difficult to interpret the origin of their performance.}

\paragraph{GPT-4 can partially solve the Korean bar exam}
Noticing all open-source LLMs and GPT-3.5 show limited performance on the bar exam, we next focus on evaluating more powerful commercial LLMs.
Five commercial LLMs--Claude-3-sonnet, Claude-3-opus, Claude-3.5-sonnet, and GPT-4--achieve higher performance compared to the strongest open-source model Qwen-72b, with improvement of +1.8, +6.8, +9.8, +11.8 on the knowledge tasks; +1.0, +0.5, +2.6, +1.9 on the two reasoning tasks (\rcausal, \rcontradict); +2.4, +9.9, +11.4, +17 on the 2024 bar exam 2024 (rows 6--8, and 10).
It shows that although GPT-4 passes the U.S. bar exam and it achieves most competent performance, 
there remains significant room for improvement in LLM applications for Korean legal AI tasks.
Here we focus on the 2024 bar exam, which is least likely to have been used in training these LLMs. 
Scores for the bar exams from 2010 to 2023, including the professional responsibility QA, are shown in Table \ref{tbl_bar_exam_criminal}, \ref{tbl_bar_exam_civil}, \ref{tbl_bar_exam_public}, \ref{tbl_bar_exam_responsibility} in Appendix.

One noticeable observation is that in the \kstatuteMatching\ task, where a model needs to match the content of a law to its statute number, all 
 models show very low performance (23.0--49.0, col 4). This indicates that current LLMs are unreliable for recalling specific legal knowledge, as previously observed \citep{fei2023lawbench,collam2024,dahl2024largelegalfiction,magesh2024lhallucinationfree}. 

\paragraph{RAG can be beneficial, but several factors influence the overall performance}
Next we evaluate GPT-4 under a RAG setting. We prepared two types of corpora, a precedent-corpus consisting of 150k Korean precedents~\citep{hwang2022lboxopen} and a statute-corpus comprising Korean statutes, and municipal ordinances and rules.
We employ a BM25 retriever without re-ranking as the baseline.
Even in this simple setting, GPT-4 achieves higher scores on the knowledge tasks, scoring +2.4 with precedents, +0.4 with statutes, and +3.3 with both precedents and statutes (rows 4--6 from the bottom, column 1).
On the bar exam, GPT-4 improves by +7.4 in \bcriminal\ and +3.3 in \bcivil\ (3rd row from the bottom, 2nd and 3rd columns from the right). 
However, in \bpublic\, GPT-4 scores -2.5 with precedents, -5.8 with statutes, and -5.8 with both, indicating a drop in performance.

Conversely, Claude series shows no clear improvement under the same RAG setting, with Claude-3-sonnet scoring -2.1, and -8.4 on the knowledge and the bar exam respectively (column 1, 14; rows 7, 16); Claude-3-opus scores -1.7, and -4.3 (column 1, 14; rows 8, 17); Claude-3.5-sonnet scores +0.1, and +3.7 (column 1, 14; rows 9, 18).
This illustrates how multiple factors--LLM, retriever, re-ranker, corpus, etc.--can influence overall performance, 
highlighting the importance of developing a benchmark to provide a foundation for the automatic evaluation of the RAG system.

\section{Analysis}
Here we provide a further analysis of the  Korean bar exam held in 2024. 
This exam was chosen because:
(1) it requires both deep legal knowledge and reasoning capabilities;
(2) GPT-4 exhibited surprisingly low performance despite having passed the U.S. bar exam \cite{martinez2023reeval_bar_exam_gpt4};
(3) GPT-4 was trained using data from before the 2024 bar exam, ruling out the possibility of data contamination.

We first categorizes individual questions into two types: (1) Rule or (2) Application.
A question is categorized as Application-type if it requires not just an understanding of legal knowledge or principles, but also how these are applied to specific real-world cases. Otherwise, it is considered Rule-type. Although the IRAC method--Issue, Rule, Application, and Conclusion (IRAC)--typically has four categories, we focus on two because questions often belong to multiple categories, making them difficult to clearly categorize.
In the Civil domain, there are 29 Rule-type questions and 41 Application-type questions. The Criminal domain includes 4 Rule-type and 36 Application-type questions, while the Public domain comprises 23 Rule-type and 17 Application-type questions.

\begin{figure}[h]
    \centering
    \includegraphics[width=0.5\textwidth]{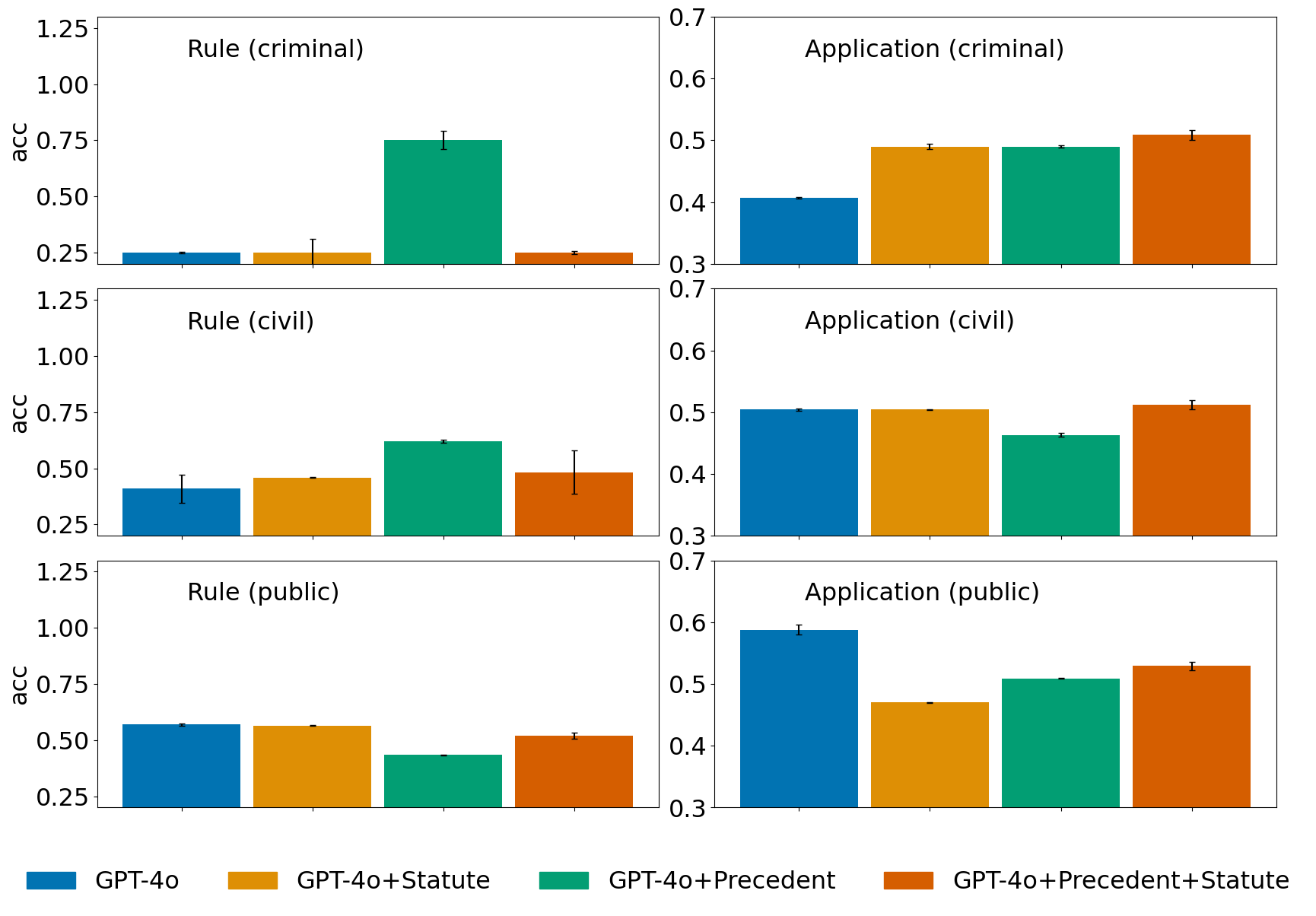}
    \caption{GPT-4's performance on the 2024 bar exam without RAG (blue) or with RAG (orange, green, red).
    }
    \label{fig: accuracy_comparison_bar_exam_RA}
\end{figure}

\paragraph{RAG primarily enhances performance on Rule-type questions}
Fig. \ref{fig: accuracy_comparison_bar_exam_RA} depicts GPT-4's accuracy on the 2024 bar exam, highlighting performance without RAG (blue), with the statute corpus (orange), with the precedent-corpus, and with both corpora (red). 
The left panels display accuracy on Rule-type questions, while the right panels show on Application-type questions. 

In both Civil and Criminal domains, using the precedent corpus significantly enhances accuracy for Rule-type questions (upper left and middle left panels, blue vs green).
However, there are no improvements in Civil Application-type questions and only marginal improvements in Application-type questions.
Conversely, in the Public domain, neither the precedent corpus nor the statute corpus proves beneficial.

\section{Conclusion}
We propose \ours, a pragmatic benchmark designed for Korean Legal language understanding that comprises (1) 7 knowledge tasks, (2) 4 reasoning tasks, and (3) Korean bar exams. The first two tasks were designed with close corporation with legal professionals and the answers were validated by lawyers. The results indicate that there is still a significant room for improvement in LLMs' capabilities in the Korean legal domain.
Additionally, recognizing that legal research often involves consulting related legal documents, 
we equip LLMs for evaluation in a RAG setting by providing two accompanying legal corpora for retrieval: a precedent corpus from a previous study~\cite{hwang2022lboxopen} and a statute corpus developed in this study. 
Enabling LLMs to utilize external legal documents via a simple BM25 retriever has shown to improve performance but not always depending on several factors. This highlights the importance of our work that provides a common playground for automatic evaluating of RAG systems. 

\section*{Limitations}
Here we evaluate LLMs on multi-choice type questions only where clear ground truths can be established.
For generative tasks, although it would be possible to use LLM-as-a-judge as a proxy, 
the field is actively evolving and its accuracy is  still limited. 
Also, it is particularly challenging to evaluate generative tasks in the legal domain  where the hallucination are still prone and where complex in-depth reasoning process are required~\cite{magesh2024lhallucinationfree}. 
Although we have meticulously designed and selected the tasks for \ours, it is important to acknowledge that our benchmark cannot encompass  the entire spectrum of legal tasks especially where the labeling data is very costly due to the fact that it requires professional trained for several years. 
Nevertheless we aim to capture the essential aspects of legal intelligence that can be automatically evaluated. 
For this, we have collaborated with legal professionals and verified all examples thoroughly, striving to establish a reliable benchmark.

\section*{Ethics Statement}
We use Korean precedents and statutes as a main source of raw data where all personal information, if any, is redacted by Korean government or court. The part of the datasets include the detailed description of crimes from precedents that are already publicly available.
Open-source LLMs have a possibility of the misuse and can be easily tuned for unethical purpose \cite{qi2024iclr_LLM_safety,kim2024openness_eve}. 
However, here we do not release or train any models but focusing on their evaluation with the benchmark that consists of legal question and answers, the precedent corpus, and the statute corpus.

\section*{Acknowledgements}
We thank Prof. Yong Lim, Prof. Jong Hwan Kim for their critical comments on the legal aspects.
We also thank Hyunjun Kim, Jinu Lee, and Dongjun Lee for their assistant in designing tasks during the early stages of the project.
The part of this research was supported by the National Research Foundation of Korea (NRF) grant funded by the Korea government (MSIT) (NO.2022M3J6A1084845) for Yeeun Kim.
\bibliographystyle{acl_natbib.bst}
\bibliography{legal_ai}

\appendix

\onecolumn
\newpage 
\section{Appendices}
\subsection{General Lessons} \label{sec: general_lessons}
While developing the benchmark for the legal domain, which requires high expertise, we learned the following lessons:
\begin{itemize}
    \item It is necessary to communicate closely with domain experts to distinguish subtle differences and accurately label the data.
    \item Freely available data, often created by individuals with semi-expertise, frequently include substantial amounts of errors (up to 21\% in our study), highlighting the importance of close collaboration with experts.
    \item The selection of domains should be done carefully, considering the difficulty of the labeling. For instance, during the creation of the Causal reasoning QA dataset, we initially selected “insurance dispute”. The domain was chosen because corresponding cases include the lexical cue “substantial causation (상당인과관계),”. However, during the verification stage by the lawyer, feedback was received that although the term “causation” might be mistaken to mean a judgment by the court, legal professionals use it in a practical sense, indicating a contextual connection between the act and the result based on common experience (경험칙). Therefore, the presence of “substantial causation” alone cannot serve as a standard to determine the causality of the case. Based on this feedback and the lawyer’s recommendation, we changed the domain to “Death or Injury Resulting from Violence” or “Death Resulting from Bodily Injury” where actions could be seen as directly and clearly causing the results (the defendant’s actions and the victim’s death or injury).
\end{itemize}

\subsection{Datasets}
\subsubsection{Legal Concept QA} \label{sec: appendix_legal_concept_qa}
Most Korean legal terms originate from Chinese characters, leading to significant differences in meaning with even small variations in characters. For instance, the Korean legal terms ‘소각하’ (so-gak-ha) and ‘소취하’ (so-chui-ha) both refer to methods of terminating a lawsuit. Despite differing by only one character, they represent significant differences regarding the party responsible for terminating the lawsuit.

Therefore, we collected terms that are rarely used by the general public but should be clearly differentiated by experts or have different meanings when used in a legal context. We first reviewed the definitions of terms listed alphabetically in a legal dictionary and prioritized terms that have subtle variations.  Additionally, during the term selection process, we used commercial LLMs (ChatGPT and Claude-sonnet) to identify 100 terms that at least one LLM responded inadequately. A lawyer then verified the dataset, identifying a single error among 100 examples, which was subsequently corrected based on the lawyer's feedback.

\newpage

\begingroup
\setlength{\tabcolsep}{4pt} 
\renewcommand{\arraystretch}{1.1} 
\scriptsize
\vspace{-\abovedisplayskip}
    \captionof{table}{Task examples. The examples are translated to English using GPT-4.}
    \label{tab:task_examples_1}
    \centering
    \begin{threeparttable}
    \begin{tabular}{|>{\raggedright\arraybackslash}p{2cm}|>{\raggedright\arraybackslash}p{5cm}|>{\raggedright\arraybackslash}p{4cm}|>{\raggedright\arraybackslash}p{2cm}|}
                    \specialrule{1.5pt}{0pt}{0pt}

        \textbf{Data Name} & \textbf{Question} & \textbf{Candidate choices} & \textbf{Answer} \\
            \specialrule{1.5pt}{0pt}{0pt}
        Legal Concept QA (\kconceptqa)
        & 다음 법률용어의 정의를 읽고 해당 단어를 선택해 주세요. 정의: 사실인정의 기초가 되는 경험적 사실을 경험자 자신이 직접 법원에 진술하지 않고 다른 형태에 의하여 간접적으로 보고하는 것을 말한다.
        & A. 직접증거 \newline B. 진술서 \newline C. 자백 \newline D. 증언보조사항 \newline E. 전문증거
        & E. 전문증거 \\
        \midrule
        & Read the definition of the following legal term and select the appropriate word. Definition: It refers to reporting an empirical fact that forms the basis for fact-finding indirectly in other forms without the experiencer directly testifying in court.
        & A. Direct Evidence \newline B. Statement \newline C. Confession \newline D. Testimony Assistance \newline E. Hearsay Evidence
        & E. Hearsay Evidence \\
        \specialrule{1.5pt}{0pt}{0pt}

        Offense Component QA (\kcomponentqa)
        & 甲은 乙을 주먹으로 때려 상해를 가하였으나, 그것은 乙이 이유 없이 甲에게 욕설을 퍼부어 행해진 것이다. 甲은 상해죄로 처벌되는가?
        & A. 아니오 \newline B. 예
        & B. 예 \\
        \midrule
        & A hit B with his fist and caused injury, but it was done because B insulted A without reason. Is A punishable for injury?
        & A. No \newline B. Yes
        & B. Yes \\
        \specialrule{1.5pt}{0pt}{0pt}
        Query Statute Matching QA (\kstatuteQueryMatching)
        & 다음 질문에 가장 관련이 깊은 법령을 선택해 주세요. 
        식품 영업을 하려면 어떤 시설기준을 충족해야되나요?
       & A. 형사소송법 제225 \newline B. 항만과그주변지역의개발및이용에관한법률 제6조 \newline C. 고용보험법 제87조 \newline D. 식품위생법 제36조1항 \newline E.변호사법 제23조
        & D. 식품위생법 제36조1항\\
        \midrule
        & Select the statute most relevant to the following question: What facility standards must be met to operate a food business?
        & A. Criminal Procedure Act Article 225 \newline B. Act on the Development and Use of Ports and Surrounding Areas Article 6 \newline C. Employment Insurance Act Article 87 \newline D. Food Sanitation Act Article 36(1) \newline E. Attorney-at-Law Act Article 23
        & D. Food Sanitation Act Article 36(1) \\
        \specialrule{1.5pt}{0pt}{0pt}
        Statute Number and Content Matching QA (\kstatuteMatching)
        & 다음 중 법률번호와 해당내용이 올바르게 연결된 것을 선택해 주세요.
        & A. 형법 제37조(상상적 경합) 한 개의 행위가 여러 개의 죄에 해당하는 경우에는 가장 무거운 죄에 대하여 정한 형으로 처벌한다. \newline
        B. 형법 제37조(피해자의 승낙) 처분할 수 있는 자의 승낙에 의하여 그 법익을 훼손한 행위는 법률에 특별한 규정이 없는 한 벌하지 아니한다. \newline
        C. 형법 제37조(구류) 구류는 1일 이상 30일 미만으로 한다. \newline
        D. 형법 제37조(경합범) 판결이 확정되지 아니한 수개의 죄 또는 금고 이상의 형에 처한 판결이 확정된 죄와 그 판결확정전에 범한 죄를 경합범으로 한다. <개정 2004. 1. 20.> \newline
        E. 형법 제37조(누범) 금고(禁錮) 이상의 형을 선고받아 그 집행이 종료되거나 면제된 후 3년 내에 금고 이상에 해당하는 죄를 지은 사람은 누범(累犯)으로 처벌한다.
        & D. 형법 제37조(경합범) 판결이 확정되지 아니한 수개의 죄 또는 금고 이상의 형에 처한 판결이 확정된 죄와 그 판결확정전에 범한 죄를 경합범으로 한다. <개정 2004. 1. 20.> \\
        \midrule
        & Choose the correct match between statute number and its content.
        & A. Criminal Act Article 37 (Ideal Concurrence) When a single act constitutes multiple crimes, it shall be punished by the heaviest penalty prescribed for such crimes. \newline
        B. Criminal Act Article 37 (Victim's Consent) An act that harms a legal interest with the consent of a person who can dispose of it is not punishable unless otherwise specified by law. \newline
        C. Criminal Act Article 37 (Detention) Detention shall be for at least one day and less than thirty days. \newline
        D. Criminal Act Article 37 (Concurrent Crimes) Multiple crimes that have not been finally adjudicated or crimes committed before the final adjudication of a sentence of imprisonment or heavier punishment shall be treated as concurrent crimes. <Amended Jan 20, 2004> \newline
        E. Criminal Act Article 37 (Repeat Offender) A person who commits a crime punishable by imprisonment or heavier punishment within three years after being released from the execution of or being exempted from a sentence of imprisonment or heavier punishment shall be punished as a repeat offender.
        & D. Criminal Act Article 37 (Concurrent Crimes) Multiple crimes that have not been finally adjudicated or crimes committed before the final adjudication of a sentence of imprisonment or heavier punishment shall be treated as concurrent crimes. <Amended Jan 20, 2004> \\
        \bottomrule
    \end{tabular}
    \end{threeparttable}
\vspace{-\belowdisplayskip}
\endgroup

\begingroup
\setlength{\tabcolsep}{4pt} 
\renewcommand{\arraystretch}{1.1} 
\begin{table*}[h]
\scriptsize
    \caption{Task examples}
    \label{tab:task_examples_2}
    \centering
    \begin{threeparttable}
    \begin{tabularx}{\textwidth}{|X|X|X|X|}

                    \specialrule{1.5pt}{0pt}{0pt}
        \textbf{Data Name} & \textbf{Question} & \textbf{Candidate choices} & \textbf{Answer} \\
            \specialrule{1.5pt}{0pt}{0pt}
        
        Statute Hallucination QA (\kstatuteHallucination)
        & 4.7m 높이의 개천 다리 밑에 그물망이 설치된 사실을 인지한 뒤 충동적으로 그곳을 향해 사람을 떨어뜨리는 행위는 살인죄나 살인미수죄가 무조건 적용되는 불법행위인가요?
        & A. 그렇다. 폭행금지법에 의해 처벌될 수 있다. \newline B. 그렇다. 살인미수금지법에 의해 처벌될 수 있다. \newline C. 그렇다. 형법에 의해 처벌된다. \newline D. 아니다. 살인미수죄로 처벌되지 않을 수 있다.
        & D. 아니다. 살인미수죄로 처벌되지 않을 수 있다. \\
        \midrule
        & After recognizing the presence of a net installed under a 4.7m high creek bridge, is the act of impulsively dropping a person toward it always considered an illegal act that automatically applies to murder or attempted murder charges?
        & A. Yes. It can be punished under the Assault Prohibition Act. \newline B. Yes. It can be punished under the Attempted Murder Prohibition Act. \newline C. Yes. It is punishable under the Criminal Act. \newline D. No. It may not be punishable as attempted murder.
        & D. No. It may not be punishable as attempted murder. \\
        \specialrule{1.5pt}{0pt}{0pt}
        Common Legal Mistake QA (\kcommonMistakeqa)
        & 화장실을 못 찾던 7살 아이가 길에서 소변을 보는 것은 불법인가요?
        & A. 정답이 없다. \newline B. 그렇다. \newline C. 아니다.
        & B. 그렇다. \\
        \midrule
        & Is it illegal for a 7-year-old child who couldn't find a bathroom to urinate on the street?
        & A. No correct answer. \newline B. Yes. \newline C. No.
        & B. Yes. \\
        \specialrule{1.5pt}{0pt}{0pt}
        Common Legal Mistake QA Reasoning (\kcommonMistakeqaR)
        & 알코올 의존증 환자가 술에 취해 살인을 저질렀다면 감형해주나요?
        & A. 그렇다 알코올 의존증 환자라면 사건 당시 술에 취했는지의 여부와 관계없이 감형될 수 있다. \newline B. 그렇다 술에 취한 사람이 범죄를 저지르면 심신미약자로 간주되어 감형될 수 있다. \newline C. 아니다 범죄가 심각하다면 범죄자가 알코올 의존증이 있다하더라도 고려되지 않는다.
        & B. 그렇다 술에 취한 사람이 범죄를 저지르면 심신미약자로 간주되어 감형될 수 있다. \\
        \midrule
        & If an alcohol-dependent person commits murder while intoxicated, is their sentence reduced?
        & A. Yes, an alcohol-dependent person can have their sentence reduced regardless of whether they were intoxicated at the time of the incident. \newline B. Yes, a person who commits a crime while intoxicated can be considered mentally impaired and have their sentence reduced. \newline C. No, if the crime is serious, it will not be considered even if the offender is alcohol-dependent.
        & B. Yes, a person who commits a crime while intoxicated can be considered mentally impaired and have their sentence reduced. \\
        \specialrule{1.5pt}{0pt}{0pt}
        Causal Reasoning QA (\rcausal)
        & 다음 [검사의 공소사실], [피고인의 주장], [증거]를 읽고 주어진 정보만을 바탕으로 질문에 답해주세요. [검사의 공소사실] [피고인의 주장] [증거] A, B를 각각 A: 피고인들의 행위 B: 피해자의 사망라고 할 때 A와 B사이의 관계를 '인과관계있음', '인과관계없음' 중 하나를 선택하여 '답변: 인과관계있음'과 같이 단답식으로 답해주세요.
        & A. 인과관계있음 \newline B. 인과관계없음
        & A. 인과관계있음 \\
        \midrule
        & Read the [prosecutor's charges], [defendant's claims], and [evidence] and answer the question based solely on the given information. [Prosecutor's charges] [Defendant's claims] [Evidence] Considering A: Defendants' actions and B: Victim's death, select 'Causal relationship exists' or 'No causal relationship' and respond with 'Answer: Causal relationship exists.'
        & A. Causal relationship exists \newline B. No causal relationship
        & A. Causal relationship exists \\
        \bottomrule
    \end{tabularx}
    \end{threeparttable}
\end{table*}
\endgroup

\begingroup
\setlength{\tabcolsep}{4pt} 
\renewcommand{\arraystretch}{1.1} 
\begin{table*}[h]
\scriptsize
    \caption{Task examples}
    \label{tab:task_examples_3}
    \centering
    \begin{threeparttable}
    \begin{tabularx}{\textwidth}{|X|X|X|X|}
                    \specialrule{1.5pt}{0pt}{0pt}
        \textbf{Data Name} & \textbf{Question} & \textbf{Candidate choices} & \textbf{Answer} \\
        \specialrule{1.5pt}{0pt}{0pt}
        Logical Contradiction QA (\rcontradict)
        & 다음 진술들이 서로 일관되는지 판단해 주세요. 진술1: 피고인이 팔꿈치로 피해자의 ‘팔뚝 부위’를 쳤다. 진술2: 피고인이 팔꿈치로 피해자의 가슴 쪽을 때렸다.
        & A. 일관되지 않음 \newline B. 일관됨
        & A. 일관되지 않음 \\
        \midrule

        & Determine whether the following statements are consistent. Statement 1: The defendant hit the victim's 'forearm area' with his elbow. Statement 2: The defendant hit the victim's chest area with his elbow.
        & A. Inconsistent \newline B. Consistent
        & A. Inconsistent \\
        \specialrule{1.5pt}{0pt}{0pt}
        Case Relevance QA query (\rcaseq)
        & 다음 판결문이 의뢰인의 주장을 뒷받침하나요? [의뢰인의 주장] 판결문 [상고인] [사실관계] [당사자들의 주장] [판사의 의견] A: 아니오, B: 예 중 하나를 선택하여 '답변: A'과 같이 단답식으로 답해주세요.
        & A. 예. \newline B. 아니오.
        & B. 아니오. \\
        \midrule
        & Does the following judgment support the client's claim? [Client's claim] Judgment [Appellant] [Facts] [Parties' claims] [Judge's opinion] Choose A: No or B: Yes and answer with 'Answer: A.'
        & A. Yes. \newline B. No.
        & B. No. \\
        \specialrule{1.5pt}{0pt}{0pt}
        Case Relevance QA President (\rcasep)
        & 다음 두 판결문은 같은 사안에 대해 다루고 있나요? [첫번째 판결문 상고인] [첫번째 판결문 사실관계] [첫번째 판결문 당사자들의 주장] [첫번째 판결문 판사의 의견] [두번째 판결문 상고인] [두번째 판결문 사실관계] [두번째 판결문 당사자들의 주장] [두번째 판결문 판사의 의견] A: 사안이 다르다, B: 사안이 다르지 않다 중 하나를 선택하여 '답변: A'과 같이 단답식으로 답해주세요.
        & A. 사안이 다르다 \newline B. 사안이 다르지 않다.
        & B. 사안이 다르지 않다. \\
        \midrule
        & Do the following two judgments address the same issue? [First judgment appellant] [First judgment facts] [First judgment parties' claims] [First judgment judge's opinion] [Second judgment appellant] [Second judgment facts] [Second judgment parties' claims] [Second judgment judge's opinion] Choose A: Different issues or B: Same issues and answer with 'Answer: A.'
        & A. Different \newline B. Same
        & B. Same  \\
        \bottomrule
    \end{tabularx}
    \end{threeparttable}
\end{table*}
\endgroup

\begingroup
\setlength{\tabcolsep}{2pt} 
\renewcommand{\arraystretch}{1} 

\begin{table*}[h!]
\scriptsize
\caption{The mean token length of the Bar exam}
\label{tbl_token_len_bar_exam}
\centering
\begin{threeparttable}
\begin{tabular}{l|c|c|c|c}
  \toprule
  \multicolumn{1}{c}{Year} &
  \multicolumn{1}{c}{\textbf{Criminal}} & 
  \multicolumn{1}{c}{\textbf{Civil}} &
  \multicolumn{1}{c}{\textbf{Public}} &
  \multicolumn{1}{c}{\textbf{Responsibility}} \\
  
  \midrule
  2010 & -& -& -& 334\\ 
  2011 &- &- &- & 365\\ 
  2012 & 468 & 470 & 554 & 358 \\
  2013 & 530 & 492 & 496 & 335\\
  2014 & 539 & 531 & 706 & 361\\
  2015 & 497 & 435 & 497 & 385\\
  2016 & 539 & 486 & 516 & 374\\
  2017 & 527 & 461 & 571 & 343\\
  2018 & 556 & 491 & 549 & 345\\
  2019 & 570 & 503 & 594 & 383\\
  2020 & 570 & 543 & 576 & 366\\
  2021 & 586 & 513 & 557 & 370\\
  2022 & 587 & 499 & 526 & 351\\
  2023 & 600 & 467 & 520 & 358\\
  2024 & 587 & 538 & 563 & - \\
  \bottomrule
\end{tabular}
\begin{tablenotes}
    \item{*} The bar exam was not administered in 2010 and 2011.
    \item{*} The professional responsibility exam for law schools has not yet been implemented as of June, 2024
\end{tablenotes}
\end{threeparttable}
\end{table*}
\endgroup

\begingroup
\setlength{\tabcolsep}{2pt} 
\renewcommand{\arraystretch}{1} 
\begin{table*}[h!]
\scriptsize
  \caption{Comparison of various models on Korean Bar Exam-criminal laws.
  }
  \label{tbl_bar_exam_criminal}
  \centering
  \begin{threeparttable}
  \begin{tabular}{l|c|cccccccccccc}
    \toprule
      & Avg.
      & 2012
      & 2013
      & 2014
      & 2015
      & 2016
      & 2017
      & 2018
      & 2019
      & 2020
      & 2021
      & 2022
      & 2023
      \\
    \midrule
    GPT-3.5
      & 20.8
      & 32.5
      & 25.0
      & 12.5
      & 20.0
      & 27.5
      & 17.5
      & 20.0
      & 17.5
      & 12.5
      & 22.5
      & 25.0
      & 22.5
      \\

    GPT-4
    & 40.4
      & 57.5 
      & 52.5 
      & 30.0 
      & 37.5 
      & 35.0 
      & 42.5 
      & 25.0 
      & 57.5 
      & 27.5 
      & 45.0 
      & 20.0 
      & 55.0 
      \\
    Claude-3-opus
    & 34.4
      & 62.5
      & 42.5
      & 25.0
      & 32.5
      & 27.5
      & 42.5
      & 12.5
      & 45.0
      & 24.0
      & 42.5
      & 17.5
      & 45.0
      \\
    \midrule
    random
    & 25 
      & 33
      & 23 
      & 28 
      & 25
      & 23 
      & 25 
      & 25
      & 20
      & 25
      & 28
      & 25 
      & 25
      \\
    \midrule
    \bottomrule
  \end{tabular}
  \end{threeparttable}
\end{table*}
\endgroup

\begingroup
\setlength{\tabcolsep}{2pt} 
\renewcommand{\arraystretch}{1} 
\begin{table*}[]
\scriptsize
  \caption{Comparison of various models on the Korean Bar Exam, Civil domain.
  }
  \label{tbl_bar_exam_civil}
  \centering
  \begin{threeparttable}
  \begin{tabular}{l|c|cccccccccccc}
    \toprule
      & Avg.
      & 2012
      & 2013
      & 2014
      & 2015
      & 2016
      & 2017
      & 2018
      & 2019
      & 2020
      & 2021
      & 2022
      & 2023
      \\
    \midrule
    GPT-3.5
      & 23.6
      & 24.3
      & 22.9
      & 21.4
      & 18.6
      & 24.3
      & 25.7
      & 24.3
      & 25.7
      & 21.4
      & 31.4
      & 14.3
      & 28.6
      \\
    GPT-4
    & 41.0
      & 41.4 
      & 35.7 
      & 52.9 
      & 40.0 
      & 34.3 
      & 50.0 
      & 20.0 
      & 42.9
      & 47.1
      & 37.1
      & 50.0
      & 40.0
      \\
    Claude-3-opus
    & 38.8
      & 37.1
      & 32.9
      & 45.7
      & 37.1
      & 35.7
      & 35.7
      & 22.9
      & 44.3
      & 40.0
      & 44.3
      & 42.9
      & 42.9
      \\
    \midrule
    random
     & 25 
      & 27 
      & 23 
      & 23
      & 23
      & 29 
      & 26 
      & 23 
      & 27 
      & 21 
      & 24 
      & 30
      & 24 
      \\
    \midrule
    \bottomrule
  \end{tabular}
  \end{threeparttable}
\end{table*}
\endgroup

\begingroup
\setlength{\tabcolsep}{2pt} 
\renewcommand{\arraystretch}{1} 
\begin{table*}[]
\scriptsize
  \caption{Comparison of various models on the Korean Bar Exam, Public domain.
  }
  \label{tbl_bar_exam_public}
  \centering
  \begin{threeparttable}
  \begin{tabular}{l|c|cccccccccccc}
    \toprule
      & Avg.
      & 2012
      & 2013
      & 2014
      & 2015
      & 2016
      & 2017
      & 2018
      & 2019
      & 2020
      & 2021
      & 2022
      & 2023
      \\
    \midrule
    GPT-3.5
      & 27.2
      & 27.5
      & 30.0
      & 27.5
      & 37.5
      & 37.5
      & 15.0
      & 25.0
      & 34.2
      & 30.0
      & 20.0
      & 30.0
      & 17.5
      \\

    GPT-4
    & 54.8
      & 60.0
      & 55.0
      & 52.5
      & 52.5
      & 65.0
      & 65.0
      & 17.5
      & 50.0
      & 57.5
      & 57.5
      & 75.0
      & 50.0
      \\
    Claude-3-opus
    & 49.2 
      & 45.0
      & 62.5
      & 52.5
      & 42.5
      & 45.0
      & 70.0
      & 22.5
      & 57.9
      & 32.5
      & 60.0
      & 52.5
      & 45.0
      \\
    \midrule
    random
     & 25 
      & 25 
      & 28 
      & 28 
      & 25
      & 23 
      & 23 
      & 23 
      & 23 
      & 25
      & 25 
      & 28 
      & 23 
      \\
    \midrule
    \bottomrule
  \end{tabular}
  \end{threeparttable}
\end{table*}
\endgroup

\begingroup
\setlength{\tabcolsep}{2pt} 
\renewcommand{\arraystretch}{1} 
\begin{table*}[]
\scriptsize
  \caption{Comparison of various models on the Korean Bar Exam, the Responsibility domain. This exam begins from 2010 following the introduction of law schools in South Korea in 2009. As of June 2024, data for the year 2024 is not included as the 15th exam is scheduled for August 2024.
  }
  \label{tbl_bar_exam_responsibility}
  \centering
  \begin{threeparttable}
  \begin{tabular}{l|c|cccccccccccccc}
    \toprule
      & Avg.
      & 2010
      & 2011
      & 2012
      & 2013
      & 2014
      & 2015
      & 2016
      & 2017
      & 2018
      & 2019
      & 2020
      & 2021
      & 2022
      & 2023
      \\
    \midrule
    GPT-3.5
      & 34.3
      & 37.5
      & 35.0
      & 45.0
      & 35.0
      & 40.0
      & 35.0
      & 40.0
      & 35.0
      & 32.5
      & 27.5
      & 27.5
      & 37.5
      & 25.0
      & 27.5
      \\

    GPT-4
    & 61.9
      & 70.0
      & 62.5
      & 77.5
      & 62.5
      & 62.5
      & 70.0
      & 57.5
      & 30.0
      & 67.5
      & 62.5
      & 52.5
      & 77.5
      & 52.5
      & 60.0
      \\
    Claude-3-opus
    & 58.9
      &  72.5
      &  60.0
      &  70.0
      &  60.0
      &  55.0
      &  57.5
      &  60.0
      &  40.0
      &  60.0
      & 67.5
      &  42.5
      & 70.0
      & 60.0
      & 50.0

      \\

    \bottomrule
  \end{tabular}
  \end{threeparttable}

\end{table*}
\endgroup

\begingroup
\setlength{\tabcolsep}{1pt} 
\renewcommand{\arraystretch}{1} 
\begin{table*}[ht!]
\scriptsize
  \caption{Comparison of various models.
  }
  \label{tbl_comp_appendix}
  \centering
  \begin{threeparttable}
  \begin{tabular}{l|c|ccccccc|c|cccc|c|ccc}
    \toprule
    \multicolumn{1}{c}{Name} &
    \multicolumn{8}{c}{\makecell{\kblk}} &
    \multicolumn{5}{c}{\makecell{\kblr}} &
    \multicolumn{4}{c}{\makecell{\kbll\ 2024$^\dagger$}} 
    \\
      & AVG$_K$  
      & \kconceptqa 
      & \kcomponentqa
      & \kstatuteMatching 
      & \kstatuteQueryMatching 
      & \kstatuteHallucination
      & \kcommonMistakeqa
      & \kcommonMistakeqaR
      & AVG$_R$  
      & \rcausal
      & \rcontradict
      & \rcaseq
      & \rcasep
      & AVG$_B$  
      & \bcriminal
      & \bcivil 
      & \bpublic 
      \\
      \midrule

      Most frequent
      & 33 
      & 20 
      & 50 
      & 20 
      & 21 
      & 25 
      & 34 
      & 35 
      & 50
      & 50
      & 50
      & 50
      & 50
      & 24
      & 23 
      & 26 
      & 23 
      \\
      \midrule
   KoGPT-0.13b$^a$ %
    & 28.7
    & 20.0
      & 50.0
      &  20.0
      &  21.2
      &  25.3
      &  31.7
      &  32.5
      & 41.1
      & 31.6
      & 50.5
      & n/a
      & n/a 
      & 16.9
      & 15.0
      & 15.7
      & 20.0
      \\
    polyglot-ko-1.3b$^b$ 
    & 28.7
    & 20.0
      & 50.0
      & 20.0
      & 21.2
      & 25.3
      & 31.7
      & 32.5
      & 48.4
      & 51.6
      & 45.1
      & n/a
      & n/a 
      & 18.6
      & 20.0
      & 15.7
      & 20.0
    \\
    polyglot-ko-12.8b$^b$ %
    & 28.7
    & 20.0
      & 50.0
      & 20.0
      & 21.2
      & 25.3
      & 31.7
      & 32.5
      & 50.6
      & 51.6
      & 49.5
      & n/a
      & n/a
      & 18.6
      & 20.0
      & 15.7
      & 20.0
    \\
    LCube-base-0.12b$^c$ %
    & 28.7
    & 20.0
      & 50.0
      & 20.0
      & 21.2
      & 25.3
      & 31.7
      & 32.5
      & 56.8
      & 68.4
      & 45.1
      & n/a
      & n/a
      & 18.6
      & 20.0
      & 15.7
      & 20.0
    \\
    \midrule 
    KORani-v3-13b (polyglot-ko)$^d$ %
    & 31.3
    & 22.0
      & 50.0
      & 20.0
      & 21.2
      & 26.7
      & 34.2
      & 45.0
      & 51.7
      & 46.3
      & 57.1
      & n/a 
      & n/a 
      & 17.7
      & 20.0
      & 15.7
      & 17.5
    \\
     Llama3-ko-8b$^e$ %
    & 28.7
    & 21.0
    & 50.0
      & 20.0
      & 21.2
      & 25.3
      & 31.7
      & 32.5
      & 56.4
      & 62.1
      & 50.6
      & n/a
      & n/a
      & 18.6
      & 20.0
      & 15.7
      & 20.0
    \\
    komt-7b (mistral-7b)$^f$ %
    & 31.6
    & 36.0
    & 42.2
      & 20.0
      & 25.0
      & 21.3
      & 36.6
      & 40.0
      & 46.9
      & 44.2
      & 49.5
      & n/a
      & n/a 
      & 22.0
      & 27.5
      & 18.6
      & 20.0
    \\
    \bottomrule
  \end{tabular}
  \begin{tablenotes}[]
  \item $^\dagger$The scores for Bar exams 2012--2023 shown in Appendix.
  \item $^a$\texttt{skt/kogpt2-base-v2}~~~$^b$\texttt{EleutherAI/polyglot-ko-1.3b} \citep{ko2023technicalreportpolyglotkoopensource}~~~$^c$\texttt{lbox/lcube-base} \citep{hwang2022lboxopen}~~~$^d$\texttt{KRAFTON/KORani-v3-13B}
  \item $^e$\texttt{beomi/Llama-3-Open-Ko-8B-Instruct-preview}~~~$^f$\texttt{davidkim205/komt-mistral-7b-v1}

  \end{tablenotes}
  \end{threeparttable}
\end{table*}
\endgroup


\begingroup
\setlength{\tabcolsep}{4pt} 
\renewcommand{\arraystretch}{1.1} 
\begin{table*}[h]
\scriptsize
\vspace{-\abovedisplayskip}
    \caption{Comparison of KMMLU and Korean Bar Exam using 60+ Fuzzy Score - Criminal Law}
    \label{tab:comparison_kmmlu_korean_bar_criminal}
    \centering
    \begin{threeparttable}
    \begin{tabularx}{\textwidth}{|X|X|X|c|}
        \toprule
        \textbf{} & \textbf{KMMLU} & \textbf{Korean Bar Exam} & \textbf{Fuzzy Score} \\
        \midrule
        Question-Answer 1
        & 주거침입죄에 관한 설명으로 옳지 않은 것은? 주거침입죄의 미수범은 처벌하지 않는다.
        & 주거침입죄에 관한 설명 중 옳지 않은 것은? (다툼이 있는 경우에는 판례에 의함) 임대차 기간이 종료된 후에는 임차인이 계속 점유하고 있는 건물에 그 소유자가 무단으로 들어가더라도 주거침입죄가 성립하지 않는다.
        & 66
        \\
        \midrule
        & Which of the following is incorrect regarding the crime of trespassing? Attempted trespassing is not punishable.
        & Which of the following statements about the crime of trespassing is incorrect? (In case of dispute, follow the precedents.) After the lease period expires, if the tenant continues to occupy the building, the owner cannot be convicted of trespassing even if they enter the building without permission.
        & 66
        \\
        \midrule
        Source
        & undisclosed (category : Law-train)
        & 변호사 시험 1회차 형사법 선택형 1책형 문제 15번
        & .
        \\
        \specialrule{0.8pt}{1pt}{1pt}
        Question-Answer 2
        & 종물에 관한 설명 중 옳지 않은 것은? (다툼이 있는 경우에는 판례에 의함) 종물은 동산이어야 하며, 부동산은 종물이 될 수 없다.
        & 종범에 관한 설명 중 옳지 않은 것을 모두 고른 것은? (다툼이 있는 경우 판례에 의함) ㄱ. 정범의 강도예비행위를 방조하였으나 정범이 실행의 착수에 이르지 못한 경우 방조자는 강도예비죄의 종범에 해당한다. ㄴ. 자기의 지휘, 감독을 받는 자를 방조하여 범죄의 결과를 발생하게 한 자는 정범에 정한 형의 장기 또는 다액에 그 2분의 1까지 가중한 형으로 처벌한다. ㄷ. 법률상 정범의 범행을 방지할 의무가 있는 자가 그 범행을 알면서도 방지하지 아니하여 범행을 용이하게 한 때에는 부작위에 의한 종범이 성립한다. ㄹ. 종범은 정범의 실행행위 중에 이를 방조하는 경우뿐만 아니라, 정범이 실행행위에 나아갔다면 실행의 착수 전에 장래의 실행행위를 예상하고 이를 용이하게 한 경우에도 종범이 성립한다. ㄱ, ㄴ 
        & 66
        \\
        \midrule
        & Which of the following is incorrect regarding accessories? (In case of dispute, follow the precedents.) Accessories must be movable property; real estate cannot be accessories.
        & Select all incorrect statements about accessories. (In case of dispute, follow the precedents.) 1. If a person assists in the preparation of a robbery but the principal does not proceed to the execution, the accessory is still guilty of attempted robbery. 2. A person who aids someone under their direction and supervision to commit a crime shall be punished with a penalty increased by up to half of the maximum or maximum fine prescribed for the principal crime. 3. If a person legally obligated to prevent a crime knowingly fails to do so and facilitates the crime, they are guilty of an accessory by omission. 4. An accessory is not only one who assists during the principal's execution but also one who facilitates the principal's future actions if the principal had proceeded with the execution. 1, 2.
        & 66
        \\
        \midrule
        Source
        & undisclosed (category : Patent) 
        &  변호사 시험 4회차 형사법 선택형 1책형 문제 4번 
        & .
        \\
        \specialrule{0.8pt}{1pt}{1pt}
        Question-Answer 3
        & 행정의 실효성 확보수단에 관한 설명으로 옳지 않은 것은?(다툼이 있는 경우 판례에 의함) 이행강제금은 형벌과 병과할 수 없다.
        & 실행의 착수시기 또는 기수시기에 관한 설명 중 옳지 않은 것은? (다툼이 있는 경우에는 판례에 의함) 부동산의 매도인이 제1차 매수인에게서 중도금을 수령한 후, 다시 제2차 매수인에게서 계약금만을 지급받더라도 배임죄의 실행의 착수는 인정된다.
        & 65
        \\
        \midrule
        & Which of the following is incorrect regarding measures to ensure administrative effectiveness? (In case of dispute, follow the precedents.) Enforcement fines cannot be combined with criminal penalties.
        & Which of the following statements about the timing of the initiation or completion of execution is incorrect? (In case of dispute, follow the precedents.) Even if the seller of real estate receives a down payment from a second buyer after receiving an installment payment from the first buyer, the initiation of execution for breach of trust is acknowledged.
        & 65
        \\
        \midrule
        Source
        & undisclosed(category : Law)
        &  변호사 시험 1회차 형사법 선택형 1책형 문제 6번
        & .
        \\
        \bottomrule
    \end{tabularx}
    \end{threeparttable}
\vspace{-\belowdisplayskip}
\end{table*}
\endgroup

\begingroup
\setlength{\tabcolsep}{4pt} 
\renewcommand{\arraystretch}{1.1} 
\begin{table*}[h]
\scriptsize
\vspace{-\abovedisplayskip}
    \caption{Comparison of KMMLU and Korean Bar Exam using 60+ Fuzzy Score - Civil Law}
    \label{tab:comparison_kmmlu_korean_bar_civil}
    \centering
    \begin{threeparttable}
    \begin{tabularx}{\textwidth}{|X|X|X|c|}
        \toprule
        \textbf{} & \textbf{KMMLU} & \textbf{Korean Bar Exam} & \textbf{Fuzzy Score} \\
        \midrule
        Question-Answer 1
        & 종물에 관한 설명 중 옳지 않은 것은? (다툼이 있는 경우에는 판례에 의함) 종물은 동산이어야 하며, 부동산은 종물이 될 수 없다.
        & 종중에 관한 설명 중 옳지 않은 것은? (다툼이 있는 경우 판례에 의함) 종중의 임원은 종중 재산의 관리·처분에 관한 사무를 처리함에 있어 종중 규약 또는 종중총회의 결의에 따라야 할 의무는 있으나 선량한 관리자로서의 주의를 다하여야 할 의무는 없다.
        & 74
        \\
        \midrule
        & Which of the following is incorrect regarding accessories? (In case of dispute, follow the precedents.) Accessories must be movable property; real estate cannot be accessories.
        & Which of the following is incorrect regarding clan associations? (In case of dispute, follow the precedents.) The officials of a clan association must follow the rules of the clan or the resolutions of the clan general meeting when managing or disposing of clan property, but they do not have a duty to act with the care of a good manager.
        & 74
        \\
        \midrule
        Source
        & undisclosed (category: Patent)
        & 변호사 시험 10회차 민사법 선택형 문제 9번
        & .
        \\
        \specialrule{0.8pt}{1pt}{1pt}
        Question-Answer 2
        & 소멸시효에 관한 설명으로 옳지 않은 것은? (다툼이 있는 경우에는 판례에 의함) 가분채무의 일부에 대한 시효이익의 포기는 허용되지 않는다.
        & 소멸시효에 관한 설명 중 옳지 않은 것은? (다툼이 있는 경우 판례에 의함) 가압류로 인한 소멸시효 중단의 효력은 가압류결정이 제3채무자에게 송달된 때에 발생하고 가압류신청 시로 소급하지 아니한다.
        & 72
        \\
        \midrule
        & Which of the following is incorrect regarding extinctive prescription? (In case of dispute, follow the precedents.) The waiver of the benefit of prescription is not allowed for part of a divisible obligation.
        & Which of the following is incorrect regarding extinctive prescription? (In case of dispute, follow the precedents.) The effect of interruption of extinctive prescription by provisional attachment occurs when the provisional attachment decision is delivered to the third debtor and does not retroact to the time of the application for provisional attachment.
        & 72
        \\
        \midrule
        Source
        & undisclosed (category: Patent)
        & 변호사 시험 9회차 민사법 선택형 1책형 문제 11번
        & .
        \\
        \specialrule{0.8pt}{1pt}{1pt}
        Question-Answer 3
        & 이행지체에 관한 설명 중 옳지 않은 것은? (다툼이 있는 경우에는 판례에 의함) 금전채무의 지연손해금채무는 금전채무의 이행지체로 인한 손해배상채무로서, 지연손해금채무가 확정된 때로부터 이행지체가 된다.
        & 이행지체에 관한 설명 중 옳은 것은? (다툼이 있는 경우에는 판례에 의함) 채무자는 확정된 지연손해금채무에 대하여 채권자의 이행청구를 받은 때로부터 지체책임을 부담하게 된다.
        & 72
        \\
        \midrule
        & Which of the following is incorrect regarding delay in performance? (In case of dispute, follow the precedents.) The obligation to pay delay damages for a monetary debt arises from the delay in the performance of the monetary debt and is in delay from the time the delay damages are confirmed.
        & Which of the following is correct regarding delay in performance? (In case of dispute, follow the precedents.) The debtor is liable for delay from the time the creditor requests performance of the confirmed delay damages.
        & 72
        \\
        \midrule
        Source
        & undisclosed (category: Patent)
        & 변호사 시험 2회차 민사법 선택형 1책형 문제 30번
        & .
        \\
        \bottomrule
    \end{tabularx}
    \end{threeparttable}

\vspace{-\belowdisplayskip}
\end{table*}
\endgroup

\begingroup
\setlength{\tabcolsep}{4pt} 
\renewcommand{\arraystretch}{1.1} 
\begin{table*}[h]
\scriptsize
\vspace{-\abovedisplayskip}
    \caption{Comparison of KMMLU and Korean Bar Exam using 60+ Fuzzy Score - Public Law (page 1)}
    \label{tab:comparison_kmmlu_korean_bar_criminal}
    \centering
    \begin{threeparttable}
    \begin{tabularx}{\textwidth}{|X|X|X|c|}
        \toprule
        \textbf{} & \textbf{KMMLU} & \textbf{Korean Bar Exam} & \textbf{Fuzzy Score} \\
        \midrule
        Question-Answer 1
        & 신뢰보호의 원칙에 대한 설명으로 옳은 것(○)과 옳지 않은 것(×)을 바르게 연결한 것은?$^*$
        & 신뢰보호의 원칙에 관한 설명 중 옳은 것(○)과 옳지 않은 것(×)을 올바르게 조합한 것은? (다툼이 있는 경우 판례에 의함) ㄱ. 당초 폐기물처리시설을 설치한다는 도시관리계획결정 및 지형도면 고시를 하였다가 폐기물처리시설 대신 광장을 설치한다는 도시관리계획 변경결정 및 지형도면 고시를 한 경우 당초 도시관리계획결정은 도시계획시설사업의 시행자 지정을 받게 된다는 공적인 견해를 표명한 것으로 볼 수 있으므로, 그 후의 도시관리계획 변경결정 및 지형도면 고시는 당초의 도시계획시설사업의 시행자로 지정받을 것을 예상하고 폐기물처리시설의 설계비용 등을 지출한 자의 신뢰이익을 침해한다. ㄴ. 행정청 내부의 사무처리준칙에 해당하는 농림사업시행지침서가 공표된 것만으로는 사업자로 선정되기를 희망하는 자가 당해 지침에 명시된 요건을 충족할 경우 사업자로 선정되어 사업자금 지원 등의 혜택을 받을 수 있다는 보호가치 있는 신뢰를 가지게 되었다고 보기 어렵다. ㄷ. 신뢰보호의 원칙은 법률이나 그 하위법규뿐만 아니라 국가관리의 입시제도와 같이 국 · 공립대학의 입시전형을 구속하여 국민의 권리에 직접 영향을 미치는 제도운영지침의 개폐에도 적용된다. ㄹ. 신뢰보호의 원칙은 행정청이 공적인 견해를 표명할 당시의 사정이 그대로 유지됨을 전제로 적용되는 것이 원칙이므로, 사후에 그와 같은 사정이 변경된 경우에는 그 공적인 견해가 더 이상 개인에게 신뢰의 대상이 된다고 보기 어려운 만큼, 특별한 사정이 없는 한 행정청이 그 견해표명에 반하는 처분을 하더라도 신뢰보호의 원칙에 위반된다고 할 수 없다.
        & 84
        \\
        \midrule
        & Which of the following correctly matches the correct (○) and incorrect (×) statements regarding the principle of protection of trust?$^*$
        & Which of the following correctly combines the correct (○) and incorrect (×) statements about the principle of protection of trust? (In case of dispute, follow the precedents.) 1. If a city management plan decision and a topographic map notification initially announced the installation of a waste treatment facility but later changed to a decision to install a plaza instead, the initial city management plan decision can be seen as an official stance that the person who spent costs on designing the waste treatment facility expected to be designated as the implementer of the city planning facility project, and thus the subsequent city management plan change and topographic map notification violate the trust interest. 2. The mere publication of the Agricultural Project Implementation Guidelines, which are internal administrative rules, does not provide sufficient grounds for an applicant to have protected trust that they will be selected as a project operator and receive benefits such as business fund support if they meet the requirements stated in the guidelines. 3. The principle of protection of trust applies not only to laws and subordinate regulations but also to the opening and closing of policy guidelines that bind the admission systems of national and public universities, which directly affect the rights of the public. 4. The principle of protection of trust generally applies based on the assumption that the circumstances at the time the administrative authority expressed its official stance remain unchanged. Therefore, if such circumstances change later, it is difficult to consider the official stance as a subject of trust for individuals. In the absence of special circumstances, the administrative authority's actions contrary to its initial stance do not necessarily violate the principle of protection of trust.
        & 84
        \\
        \midrule
        Source
        & undisclosed (category: Law)
        & 변호사 시험 10회차 공법 선택형 문제 21번
        & .
        \\
        \bottomrule
    \end{tabularx}
    \end{threeparttable}
      \begin{tablenotes}[]
        \item         $^*$ It appears that part of the question is missing in the KMMLU dataset.

    \end{tablenotes}

\vspace{-\belowdisplayskip}
\end{table*}
\endgroup

\begingroup
\setlength{\tabcolsep}{4pt} 
\renewcommand{\arraystretch}{1.1} 
\begin{table*}[h]
\scriptsize
\vspace{-\abovedisplayskip}
    \caption{Comparison of KMMLU and Korean Bar Exam using 60+ Fuzzy Score - Public Law (page 2)}
    \label{tab:comparison_kmmlu_korean_bar_criminal}
    \centering
    \begin{threeparttable}
    \begin{tabularx}{\textwidth}{|X|X|X|c|}
        \toprule
        \textbf{} & \textbf{KMMLU} & \textbf{Korean Bar Exam} & \textbf{Fuzzy Score} \\
        \midrule
        Question-Answer 2
        & 행정심판의 재결의 기속력에 대한 설명으로 옳지 않은 것은? A: 재결이 확정된 경우에는 처분의 기초가 된 사실관계나 법률적 판단이 확정되고 당사자들이나 법원은 이에 기속되어 모순되는 주장이나 판단을 할 수 없게 된다. B: 재결에 의하여 취소되거나 무효 또는 부존재로 확인되는 처분이 당사자의 신청을 거부하는 것을 내용으로 하는 경우에는 그 처분을 한 행정청은 재결의 취지에 따라 다시 이전의 신청에 대한 처분을 하여야 한다. C: 재결의 기속력은 재결의 주문 및 그 전제가 된 요건 사실의 인정과 판단에 대하여만 미친다. D: 당사자의 신청을 받아들이지 않은 거부 처분이 재결에서 취소된 경우, 그 재결의 취지에 따라 이전의 신청에 대하여 다시 어떠한 처분을 하여야 할지는 처분을 할 때의 법령과 사실을 기준으로 판단하여야 하므로, 행정청은 종전 거부 처분 또는 재결 후에 발생한 새로운 사유를 내세워 다시 거부 처분을 할 수 있다.
        & 행정행위의 효력에 관한 설명 중 옳지 않은 것은? (다툼이 있는 경우 판례에 의함) A: 민사소송에 있어서 어느 행정처분의 당연무효 여부가 선결문제로 되는 때에는 행정처분에 당연무효 사유가 있는지 여부를 판단하여 당연무효임을 전제로 판결할 수 있고 반드시 행정소송 등의 절차에 의하여 그 취소나 무효확인을 받아야 하는 것은 아니다. B: 행정처분이 불복기간의 경과로 확정된 경우에는 그 처분의 기초가 된 사실관계나 법률적 판단이 확정되고 당사자들이나 법원이 이에 기속되어 모순되는 주장이나 판단을 할 수 없다. C: 과세처분에 관한 이의신청절차에서 과세관청이 이의신청 사유가 옳다고 인정하여 과세처분을 직권으로 취소한 이상 그 후 특별한 사유 없이 이를 번복하고 종전 처분을 되풀이하는 것은 허용되지 않는다. D: 과세처분에 대한 쟁송이 진행 중에 과세관청이 그 과세처분의 납부고지 절차상의 하자를 발견한 경우에는 위 과세처분을 취소하고 절차상의 하자를 보완하여 다시 동일한 내용의 과세처분을 할 수 있고, 이와 같은 새로운 처분이 행정행위의 불가쟁력이나 불가변력에 저촉되는 것은 아니다. E: 형사법원이 판결을 내리기 전에 영업허가취소처분이 행정쟁송절차에 의하여 취소되었다면, 그 영업허가취소처분 후의 영업행위는 무허가행위가 아닌 것이 되므로 형사법원은 그 영업허가취소처분 후의 영업행위에 대해 무죄를 선고하여야 한다.
        & 79
        \\
        \midrule
        & Which of the following is incorrect regarding the binding force of administrative appeal decisions? A: Once a decision is finalized, the factual and legal determinations underlying the decision are established, and the parties and courts are bound by them, precluding contradictory arguments or judgments. B: If a decision cancels or confirms the nullity or non-existence of a disposition that denied the applicant's request, the administrative authority must reprocess the original application in accordance with the decision. C: The binding force of a decision extends only to the ruling of the decision and the recognition and judgment of the prerequisite facts. D: If a rejection disposition is canceled by a decision, the administrative authority must reprocess the original application based on the laws and facts at the time of the disposition, allowing the authority to issue a new rejection based on new reasons arising after the original rejection or decision.
        & Which of the following is incorrect regarding the effects of administrative acts? (In case of dispute, follow the precedents.) A: In civil litigation, if the nullity of an administrative disposition becomes a preliminary issue, the court can determine its nullity without requiring the cancellation or confirmation of nullity through administrative litigation procedures. B: Once an administrative disposition is finalized due to the lapse of the appeal period, the factual and legal determinations underlying the disposition are established, and the parties and courts are bound by them, precluding contradictory arguments or judgments. C: If the tax authority cancels a tax disposition ex officio during an objection procedure due to the validity of the objection reasons, it cannot, without special reason, reverse this and repeat the previous disposition. D: During a tax dispute, if the tax authority finds procedural defects in the tax notice, it can cancel the defective tax disposition, correct the procedural defect, and reissue the same tax disposition, which does not violate the non-disputability or irrevocability of administrative acts. E: If an administrative cancellation of a business permit is reversed through administrative litigation before a criminal court's ruling, any business activity conducted after the cancellation is not considered unlicensed, and the criminal court must acquit the defendant of conducting business without a permit.
        & 79
        \\
        \bottomrule
    \end{tabularx}
    \end{threeparttable}
\vspace{-\belowdisplayskip}
\end{table*}
\endgroup

\begingroup
\setlength{\tabcolsep}{4pt} 
\renewcommand{\arraystretch}{1.1} 
\begin{table*}[h]
\scriptsize
\vspace{-\abovedisplayskip}
    \caption{Comparison of KMMLU and Korean Bar Exam using 60+ Fuzzy Score - Public Law (page 3)}
    \label{tab:comparison_kmmlu_korean_bar_criminal}
    \centering
    \begin{threeparttable}
    \begin{tabularx}{\textwidth}{|X|X|X|c|}
        \toprule
        \textbf{} & \textbf{KMMLU} & \textbf{Korean Bar Exam} & \textbf{Fuzzy Score} \\
        \midrule
        Question-Answer 3
        & 행정의 실효성 확보수단에 관한 설명으로 옳지 않은 것은?(다툼이 있는 경우 판례에 의함) 이행강제금은 형벌과 병과할 수 없다.
        & 甲은 2023. 1.경 도로에서 운전면허를 받지 아니하고 혈중알코올농도 0.15\%의 술에 취한 상태에서 자동차를 운전하였다. 검사는 甲에 대하여 무면허운전의 점에 관하여만 도로교통법위반(무면허운전)죄로 공소를 제기하였는데, 제1심 제1회 공판기일에 이르러 음주운전의 점에 관한 도로교통법위반(음주운전)죄를 추가하는 취지의 공소장변경허가신청서를 제출하였다. 이에 관한 설명 중 옳은 것을 모두 고른 것은? (다툼이 있는 경우 판례에 의함) ㄱ. 甲에 대한 도로교통법위반(무면허운전)죄와 도로교통법위반(음주운전)죄는 상상적 경합관계에 있다. ㄴ. 만약 甲이 운전한 장소가 ｢도로교통법｣상 도로가 아니라면, 도로교통법위반(무면허운전)죄는 성립할 수 있지만 도로교통법위반(음주운전)죄는 성립할 수 없다. ㄷ. 제1심법원이 공소장변경허가신청에 대한 결정을 공판정에서 고지한 경우, 그 사실은 공판조서의 필요적 기재사항이다. ㄹ. 제1심법원이 공소장변경허가신청에 대하여 불허가 결정을 한 경우, 검사는 이에 불복하여 그 결정에 대한 즉시항고를 제기할 수 있다. ㄱ, ㄷ
        & 79
        \\
        \midrule
        & Which of the following is incorrect regarding measures to ensure administrative effectiveness? (In case of dispute, follow the precedents.) Enforcement fines cannot be combined with criminal penalties.
        & In January 2023, A drove a car on a road without a driver's license and while his blood alcohol concentration was 0.15\%. The prosecutor indicted A for violating the Road Traffic Act (unlicensed driving) solely for unlicensed driving. At the first trial session, the prosecutor submitted an application to amend the indictment to add the charge of violating the Road Traffic Act (drunk driving). Which of the following statements is correct? (In case of dispute, follow the precedents.) 1. The crimes of violating the Road Traffic Act (unlicensed driving) and violating the Road Traffic Act (drunk driving) concerning A are in an imaginary concurrence relationship. 2. If the place where A drove was not a road under the Road Traffic Act, the crime of violating the Road Traffic Act (unlicensed driving) could be established, but the crime of violating the Road Traffic Act (drunk driving) could not be established. 3. If the first trial court announces the decision to permit the amendment of the indictment in the courtroom, this fact must be recorded in the trial transcript. 4. If the first trial court denies the application to amend the indictment, the prosecutor can immediately appeal the decision. 1, 3.
        & 79
        \\
        \midrule
        Source
        & undisclosed (category: Law)
        & 변호사 시험 13회차 형사법 선택형 문제 34번
        & .
        \\
        \bottomrule
    \end{tabularx}
    \end{threeparttable}
\vspace{-\belowdisplayskip}
\end{table*}
\endgroup

\end{document}